\documentclass[journal]{IEEEtran}
\usepackage{epsfig,latexsym}
\usepackage{float}
\usepackage{indentfirst}
\usepackage{amsmath}
\usepackage{amssymb}
\usepackage{times}
\usepackage{psfrag}
\usepackage{cite}
\usepackage{textcomp}
\usepackage{multirow}
\usepackage{multicol}
\usepackage{stfloats}
\usepackage{url}
\usepackage{subfigure}
\usepackage{multirow}
\usepackage{booktabs}
\usepackage{graphicx}
\usepackage{bm}
\usepackage{algorithm}
\usepackage{algorithmicx}
\usepackage{algpseudocode}

\usepackage{hyperref}
\hypersetup{hypertex=true,
	colorlinks=true,
	linkcolor=red,
	anchorcolor=green,
	citecolor=green}

\usepackage{color}
\usepackage{mathrsfs}
\usepackage{listings}
\usepackage{comment}
\usepackage{amsfonts,amssymb} 
\ifCLASSINFOpdf
\else
   \usepackage[dvips]{graphicx}
\fi
\usepackage{url}

\usepackage{graphicx}

\begin{document}
\title{ Continuous Space-Time Video Super-Resolution Utilizing Long-Range Temporal Information }

\author{Yuantong Zhang, Daiqin Yang, 
	Zhenzhong Chen,~\IEEEmembership{Senior~Member,~IEEE},
	and Wenpeng Ding		
	\thanks{This work is supported in part by Media Innovation Lab, Architecture and Technology Innovation Department, Cloud BU, Huawei.  Y. Zhang, D. Yang and Z. Chen are with the school of Remote Sensing and Information Engineering, Wuhan University, Hubei 430079, China. W. Ding is with Media Innovation Lab, Architecture and Technology Innovation Department, Cloud BU, Huawei. (Corresponding author: Zhenzhong Chen, e-mail: zzchen@ieee.org)
	}
}

\maketitle

\begin{abstract}
In this paper, we consider the task of space-time video super-resolution (ST-VSR), namely, expanding a given source video to a higher frame rate and resolution simultaneously. However, most existing schemes either consider a fixed intermediate time and scale in the training stage or only accept a preset number of input frames (e.g., two adjacent frames) that fails to exploit long-range temporal information. To address these problems, we propose a continuous ST-VSR (C-STVSR)  method that can convert the given video to any frame rate and spatial resolution. To achieve time-arbitrary interpolation, we propose a forward warping guided frame synthesis module and an optical-flow-guided context consistency loss to better approximate extreme motion and preserve similar structures among input and prediction frames. In addition, we design a memory-friendly cascading depth-to-space module to realize continuous spatial upsampling. Meanwhile, with the sophisticated reorganization of optical flow, the proposed method is memory friendly, making it possible to propagate information from long-range neighboring frames and achieve better reconstruction quality. Extensive experiments show that the proposed algorithm has good flexibility and achieves better performance on various datasets compared with the state-of-the-art methods in both objective evaluations and subjective visual effects. 
\end{abstract}


\IEEEpeerreviewmaketitle

\section{Introduction}

\IEEEPARstart{W}{ITH} the rapid development of multimedia technology,
High-resolution (HR) slow-motion video sequences are also becoming more popular since they can provide more visually appealing details. however, when we record a video with
a camera or smartphone, it is often stored with limited
spatial resolutions and temporal frame rates. Space-Time video super-resolution (ST-VSR) aims to convert low-frame-rate and low-resolution videos to higher spatial and temporal resolutions, which finds  a wide range of applications\cite{su2017deep,flynn2016deepstereo}.

  With the exponential increase of multimedia big data, deep neural network approaches have shown great advantages in various video restoration tasks. In fact, space-time video super-resolution can be divided into
two sub-tasks: Video frame interpolation (VFI) and video super  resolution (VSR). In reality, one may adopt a VFI and a VSR method separately to realize spatio-temporal upscaling for a given video.
However, a video sequence's temporal and spatial information is strongly correlated. Two-stage methods will not only introduce additional computational complexity but inevitably fail to explore the spatio-temporal relationship thoroughly. 

Rather than performing VFI and VSR independently, Researchers have recently begun to favor modeling two tasks as a joint task\cite{2021Zooming,xu2021temporal} to better exploit the more efficient space-time representation.

Although existing ST-VSR work has made significant progress, some problems remain unsolved, which hinder the practical application of space-time super-resolution technology. One of the most critical points is that most present methods are not flexible. Given a frame sequence,  most algorithms can only expand it to pre-defined intermediate moments or resolutions. Although some methods have tried to solve this problem to some extent, they still suffer from temporal inconsistency and lack the capability of exploiting long-range temporal information. Specifically, TMNet\cite{xu2021temporal} implements a temporal modulation network to interpolate arbitrary intermediate frame(s), but this kernel-based motion estimation method often results in serious temporal inconsistencies when dealing with large motions. The work most relevant to our approach is USTVSR\cite{shi2021learning} and VideoINR\cite{chen2022videoinr}. Both these two methods can modulate the input video to arbitrary resolution and frame rate, but they only consider the LR input within two neighboring video frames, and information from a long distance is ignored, which severely limiting their performance. At the same time, these methods are prone to suffer from severe blurring when dealing with extreme motions and fail to synthesize stable and continuous temporal trajectories, bringing about severe degradation in subjective visual quality.
To address these problems, we propose a continuous space-time video super-resolution method. It provides excellent spatiotemporal flexibility and integrates a powerful ability to handle large motions, which allows for synthesizing more visually appealing results.
Our contributions are summarized as follows:
 \begin{itemize}
 	\item  A continuous ST-VSR framework is proposed, which incorporates a time-arbitrary interpolation module and a scale-arbitrary upsampling module to achieve arbitrary spatio-temporal super-resolution.
 	\item  For arbitrary temporal interpolation, a lightweight frame interpolation module and a novel flow-guided texture consistency loss is proposed. Benefiting from the straightforward yet effective holes filling design and the optical flow reorganization trick, the proposed model can deal with extreme motions better meanwhile maintaining high efficiency. 
 	\item For arbitrary spatial upsampling, a simple yet effective cascading depth-to-space module is designed. We discard complex scale-aware pixel adaptation modules for memory-saving, such that it is possible to deal with long input sequences and aggregate long-term temporal information. 
\end{itemize}
By tapping into long-term space-time knowledge with a memory-friendly design, our method outperforms existing methods in both objective quality and subjective visual effects.
The remainder of the paper is organized as follows. In Section II, we review some related work. Section III presents the details of the proposed method. Experimental results and analysis are given in Section IV. Finally, the paper is concluded in Section V.

\section{related Work}
\label{sec:intro}
Our work is mainly related to three video enhancement tasks, video super-resolution, video frame interpolation, and space-time video super-resolution.
\smallskip
\subsection{Video Frame Interpolation}
Video Frame Interpolation (VFI) targets synthesizing in-between frames with their adjacent reference data.  Recent learning-based VFI approaches can be roughly divided into two categories: optical-flow-based methods\cite{0Super,DAIN,2020Softmax} and kernel-based methods\cite{2017Video,2020Video}. For flow-based methods, Bao \emph{et al.}  \cite{DAIN}  develops a depth-aware flow projection layer to synthesize intermediate frames.  SoftSplat\cite{2020Softmax} first calculates the bidirectional flows between input frames and then forward warp the corresponding feature maps via softmax splatting. For kernel-based methods, Niklaus \emph{et al.} propose several  algorithms\cite{2017Video,niklaus2021revisiting} and use local convolution to hallucinate pixels for the target frame. 
  Choi \emph{et al.}\cite{choi2021affine} propose an affine transformation-based deep frame prediction framework and integrate it in the HEVC\cite{HEVC} video coding pipeline to replace the traditional bi-directional prediction methods. Most recently, Zhou \emph{et al.}\cite{zhou2022exploring} propose a kernel-based cross-scale alignment module and a plug-in color consistency loss which is capable of improving the performance of existing VFI
frameworks.

\subsection{Video Super-Resolution}
Video super-resolution (VSR) aims to reconstruct a series of
high-resolution frames from their low-resolution
counterparts. Unlike single image super-resolution(SISR), the key to video super-resolution is mining the information between multiple frames. The recent methods can be divided into two categories: temporal sliding window based methods\cite{caballero2017real,2019EDVR,wen2022video}
and iterative-based methods\cite{2018Deep,2020VideoRSDN,sajjadi2018frame,2019Recurrent}.
For sliding window based methods, Wang\cite{wang2020deep} \emph{et al.} propose an end-to-end VSR network to super-resolve
both optical flows and images so as to recover finer details. 
EDVR \cite{2019EDVR}, as a representative method, uses DCNs\cite{2017Deformable} in a 
multi-scale pyramid and utilizes multiple attention layers to adopt alignment and integrate the features. Wen \emph{et al.} propose an end-to-end network that dynamically generates the
spatially adaptive filters for the alignment, which are constituted
by the local spatio-temporal channels of each pixel to avoid  explicit motion compensation.

For iterative-based methods, Tao \emph{el al}. \cite{tao2017detail} propose a sub-pixel motion compensation layer in a CNN framework and utilize ConvLSTM \cite{2015Convolutional} module for capturing long-range temporal information. 

Recently, BasicVSR series\cite{2020BasicVSR,chan2021basicvsr++} and some other work\cite{yi2021omniscient} stress that it is essential to utilize both neighboring LR frames and long-distance LR frames (previous and subsequent) to reconstruct HR frames.

\subsection{Space-Time Video Super-Resolution}
Space-Time Video Super-Resolution (ST-VSR) aims to transform a low spatial resolution video with 
a low frame rate into a video with higher spatial and temporal resolutions. Recently, Some deep learning-based work\cite{2020Space,2021Zooming,xu2021temporal} has made great progress in both speed and effect. Zooming Slow-Mo \cite{2021Zooming} develops a unified framework with deformable ConvLSTM to align and aggregate temporal information before performing feature fusion for ST-VSR. Based on Zooming Slow-Mo, Xu \emph{et al.} \cite{xu2021temporal} propose a temporal modulation network for controllable feature interpolation, which can interpolate arbitrary intermediate frames. Wang \emph{et al.}\cite{wang2022bi} propose a bidirectional recurrent space-time upsampling network to utilize auxiliary information at various time steps for one-stage ST-VSR.  Most recently, Chen \emph{et al.} \cite{chen2022videoinr} construct an implicit neural network to model the continuous representation of a given video. The learned implicit representation can be decoded to HR results of arbitrary spatial and temporal resolutions. Despite the remarkable progress of the aforementioned work, these efforts only consider information within a short temporal range, and their ability to handle large movements and learn from long-term sequences is severely limited.

\section{Methods}

\begin{figure*}[htbp]
    \centering
    \includegraphics[width=15.5cm]{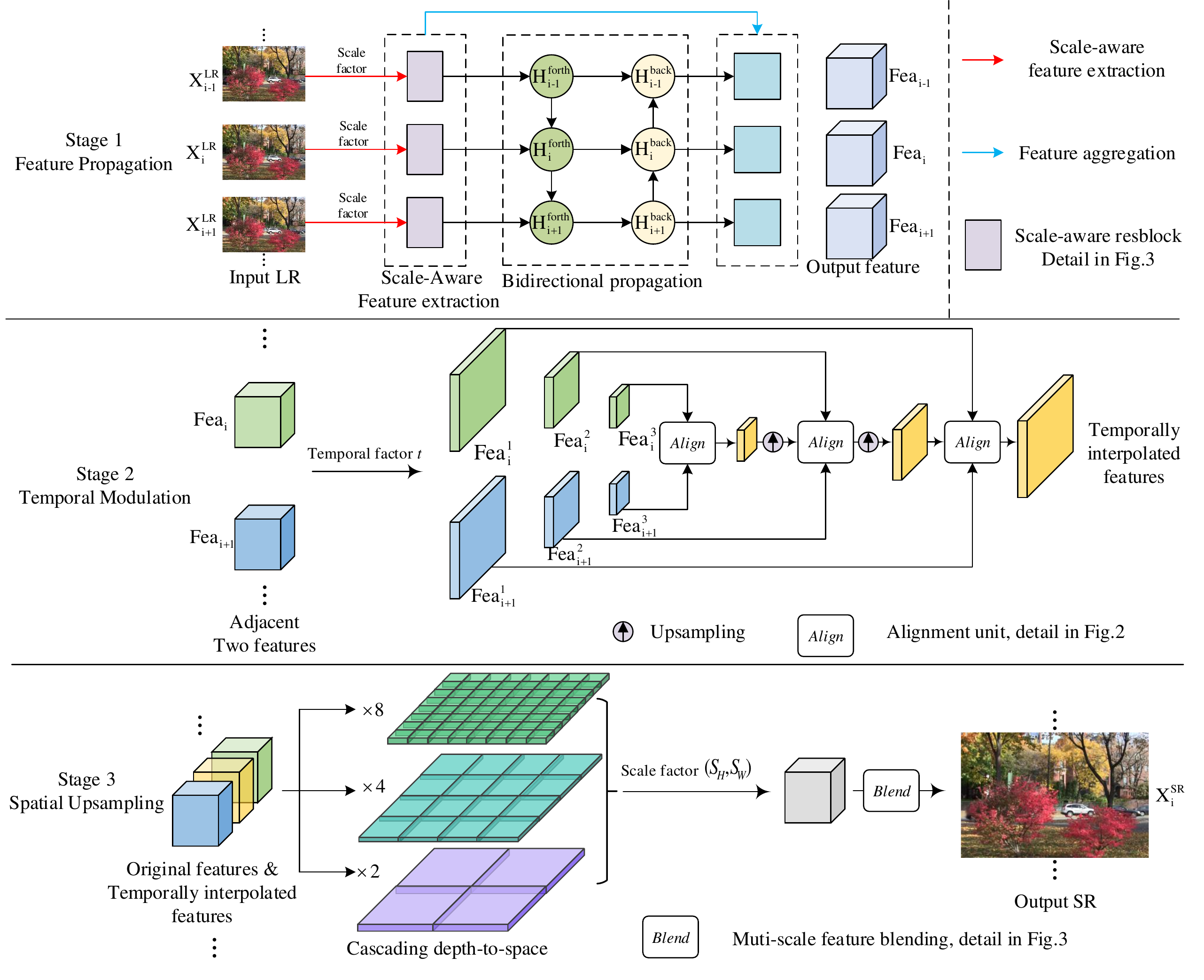}
    \caption{The framework of the proposed method, which can be divided into three stages. The input features first go through a bidirectional RNN network for feature propagation to achieve multiple-frame feature enhancement. Afterward, we synthesize the latent representation of intermediate states with two adjacent temporal features. Finally, the features obtained in the first two steps are fed into an upsampling module to achieve scale-arbitrary super-resolution. } \label{fig:1}
\end{figure*}
Given a sequence of LR inputs with $N$ frames, $I_{LR} \in \mathbb{R}^{N\times H \times W}$ also the desired temporal  upscaling factor $R$ and spatial upscaling factor $(S_{H},S_{W})$, our goal is to learn a model $\mathcal{\phi}$ to restore the corresponding high-resolution and high-frame-rate video sequence:
\begin{equation}
    \begin{aligned}
       I_{SR} &= \mathcal{\phi}(I_{LR},R,S_{H},S_{W}) ,I_{SR} \in \mathbb{R}^{R*N\times H'\times W'} ,\\
      & where \quad H' = H*S_{H},W' = W*S_{W}.
    \end{aligned}
    \label{f8}
\end{equation}
For ease of understanding,  we first give an overview of the proposed method before diving into the specific method. The details of each module will be discussed in the corresponding section.
 \subsection{ Basic Structure}\label{basic}
 As shown in Fig~\ref{fig:1}, our method consists of three seamless stages:
 Feature propagation, Temporal modulation, and Spatial upsampling.\\
 \textbf{Feature propagation} 
In the first stage, we follow the bidirectional recurrent structure from \cite{2021Zooming,2020BasicVSR,yi2021omniscient} to propagate information between input frames. Precisely, the model consists of two sub-branches, the forward and backward branches,  which propagate information through continuously updated hidden states. A skip-connection structure is also introduced to facilitate the aggregation of features at different depths. After the bidirectional propagation, the information from the entire input sequence is well exploited for subsequent reconstruction. Specifically, we use flow-guided deformable alignment\cite{chan2021basicvsr++} to aggregate information in the bidirectional propagation process and employ a pre-trained Spynet\cite{ranjan2017optical} to generate optical flow across frames.\\
  \textbf{Temporal modulation}
In the second stage, our goal is to interpolate the desired intermediate features using the two adjacent pre-existing features generated in stage one. We propose a cross-scale forward warping guided alignment module(FWGA) to approximate the intermediate feature. Through experiments, we find that  FWGA can produce more plausible results when dealing with extreme motions with high efficiency. In addition, we propose a flow-guided texture consistency loss, which can further improve the quality of interpolated frames while significantly reducing the computational cost of patch matching. \\
  \textbf{Spatial upsampling}
  In the last stage, we aim to generate the final output SR for pre-existing features achieved in stage 1 and temporal interpolated features in stage 2.
We present a memory-saving cascading depth-to-space module to realize scale-arbitrary upsampling. The proposed method's advantages and differences from the existing single-image super-resolution(SISR) are also discussed in this section.

 \subsection{ Temporal  Modulation }
  \begin{figure*}[htbp]
    \centering
    \includegraphics[width=16cm]{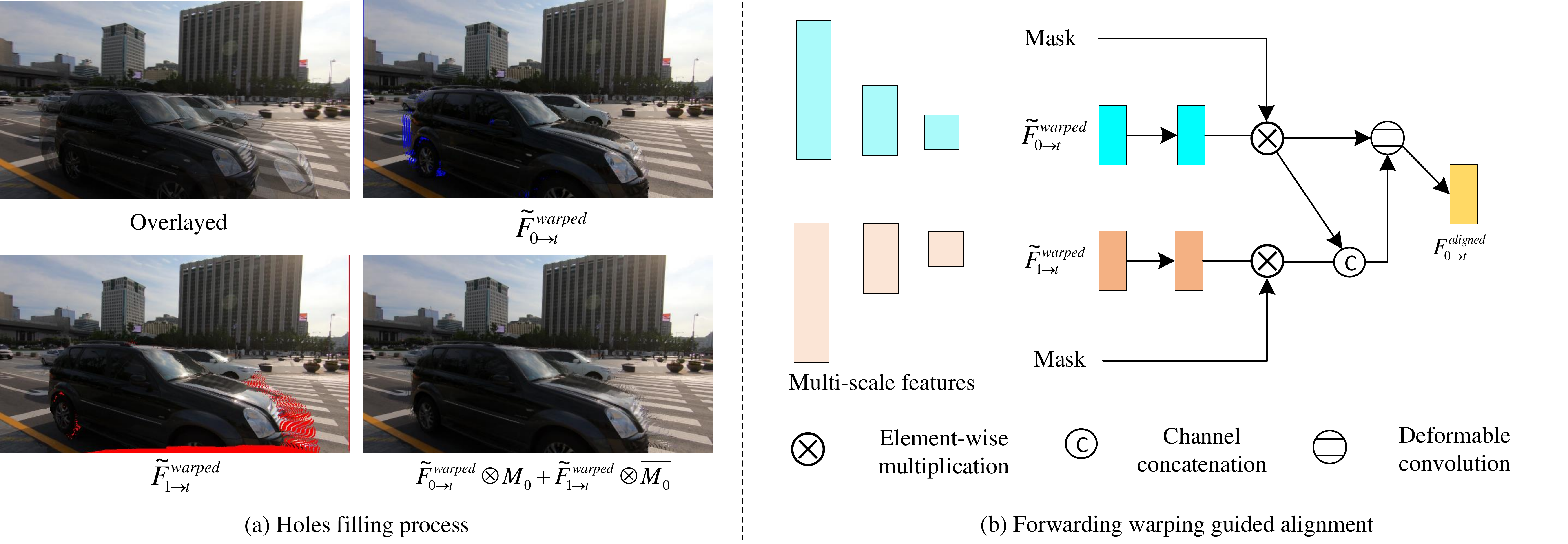}
    \caption{In (a), we show a series of images including the forward warping results from  $I_{0}$ to $I_{t}$ and $I_{1}$ to $I_{t}$.
    	 For better illustration, we colored the holes introduced by forward warping with \textcolor{blue}{blue} and \textcolor{red}{red}, respectively. In (b), we synthesize the intermediate feature in a coarse-to-fine manner, that is: Forward warping is performed first and deformable convolution is further introduced to approximate finer details.     } \label{fig:2}
\end{figure*}
After the feature representations are obtained in the first stage, we proceed to interpolate intermediate temporal features in the second stage.
In fact, many VFI methods\cite{2020Softmax,0Super,sim2021xvfi
} can achieve arbitrary intermediate frame synthesis using the optical flow technique. Our work differs from previous work in two ways. First, most current VFI methods aim to approximate an intermediate frame with adjacent two\cite{0Super,sim2021xvfi} or four\cite{xu2019quadratic,kalluri2021flavr} frames. Since for vanilla video frame interpolation, the information from too far frames may not be so helpful for synthesizing the current frame. However, we found that long-range temporal information is critical for space-time video super-resolution and aggregating long-term information leads to higher performance. Second, to handle large motion and occlusion, researchers tend to design more and more complex motion compensation\cite{2020Softmax} and post-processing modules\cite{park2021asymmetric}, which are very time and memory-consuming. Adopting such complex VFI modules for every synthetic intermediate frame in a recurrent structure is impractical. Therefore, a lightweight and efficient frame interpolation module is needed.

The flowchart of the proposed method is illustrated in Fig.~\ref{fig:1}, stage2, and the details of the alignment unit are given in Fig.~\ref{fig:2}. This module is based on our previous work\cite{zhangVCIP}. Compared to our previous work, we have improved its performance through multi-scale feature alignment and a flow-guided texture consistency loss which will be detailed in ~\ref{FGloss}.
 Inspired by EDVR\cite{2019EDVR}, the interpolation process is applied in a  multi-scale aggregation manner to better address the complex and variable motions. We first use stride convolution to obtain three-level features and perform a coarse-to-fine alignment strategy. Specifically, given two neighboring features $F_{0,1}$ that contain information from preceding and subsequent states,  our goal is to interpolate the intermediate features corresponding to the anchor moment $t \in (0,1)$. This process can be described as follows:
\begin{equation}
    \begin{aligned}
        \widetilde{F}^{l}_{0 \rightarrow t} &= t \cdot \mathcal{FW}(F^{l}_{0},\mathcal{V}_{0 \rightarrow 1}) ,\\
        \widetilde{F}^{l}_{1 \rightarrow t} &= (1-t) \cdot \mathcal{FW}(F^{l}_{1},\mathcal{V}_{1 \rightarrow 0}) ,
    \end{aligned}
    \label{f9}
\end{equation}
where  $F^{l}_{\{0,1\}}$ denotes two adjacent features at level $l$ and $\mathcal{FW}$ denotes forward warping. In general, forward warping suffers from hole issues and pixel confliction problems when multiple source pixels move to the same target pixel.
We adopt softmax splatting\cite{2020Softmax} to resolve the pixel confliction problem. Regarding holes issues, different from \cite{2020Softmax,2018Context},  we use a very light module to refill the missing context. Following the assumption of \cite{0Super}, if a pixel is visible at the moment $t$ it is most likely at least visible in one of the two successive images. We also observe that for either of the two warped frames, the majority of holes are complementary to their counterparts( That is to say, in Fig.~\ref{fig:2} (a), most \textcolor{red}{red} areas and \textcolor{blue}{blue} areas are non-overlapping ). So we use two masks to reveal these regions and complement them with pixels in their corresponding positions. This process can be formulated as follows: 
\begin{equation}
    \begin{aligned}
        &F^{l}_{0 \rightarrow t} =   \widetilde{F}^{l}_{0 \rightarrow t} \otimes \mathcal{M}^{l}_{0}+\widetilde{F}^{l}_{1 \rightarrow t} \otimes \overline{\mathcal{M}}^{l}_{0}, \\
         &F^{l}_{1 \rightarrow t} =   \widetilde{F}^{l}_{1 \rightarrow t} \otimes \mathcal{M}^{l}_{1}+\widetilde{F}^{l}_{0 \rightarrow t} \otimes \overline{\mathcal{M}}^{l}_{1},  \\
        with \quad & \mathcal{M}^{l}_{\{0,1\}}(x,y) = 
         \begin{cases}
            0 & \text{ $if \quad  F^{l}_{\{0,1\} \rightarrow t}(x,y) = 0  $ } \\
            1 & \text{ $ otherwise ,$ }
        \end{cases}
    \end{aligned}
    \label{f10}
\end{equation}
Here, we define two binary masks $\mathcal{M}^{l}_{\{0,1\}}$ for the two warped features, pixels are labeled as 0 for holes 
and 1 otherwise, and $\overline{\mathcal{M}}$ refers to elementwise-wise NOT. Please note that, in Fig.~\ref{fig:2}, we show the holes filling method with color images for better visualization. This process is actually performed in the feature space in the network.
In reality, this simple process can achieve seemingly plausible results in many cases.
However, some misaligned pixel points still exist due to inaccurate optical flow estimation and complex non-linear motions in the real world. To relax the overly strict restriction of such alignment by optical flow warping, we further introduced DCN\cite{2017Deformable} to hallucinate the intermediate frame so that the network could further correct for misalignment through learning. The whole process of multi-scale feature interpolation is carried out from low resolution to high resolution. The low-resolution results are used to guide the high-resolution alignment, which can be formulated as follows:
  \begin{equation}
    \begin{aligned}
       \Delta{P}_{0 \rightarrow t}^{l} &= f([F^{l}_{0 \rightarrow t},F^{l}_{1 \rightarrow t}], (\Delta{P}^{l+1})^{\uparrow 2}), \\
        (F^{aligned}_{0 \rightarrow t})^{l} &= DConv(F^{l}_{0 \rightarrow t},\Delta{P}_{0 \rightarrow t}^{l}),
    \end{aligned}
    \label{f11}
\end{equation}
where $f$ denotes several convolution layers to estimate the offsets, $\Delta{P}_{0 \rightarrow t}^{l}$ refers to the estimated offset at level $l$, and ${( \dots)}^{\uparrow s}$ represents  upscaling by a factor $s$.
$DConv$ denotes deformable convolution\cite{2017Deformable}.
Finally, features from both sides are blended by a learnable weight $\mathcal{W}$ to get the intermediate feature $F_{t}$ at moment $t$:
  \begin{equation}
    \begin{aligned}
        \mathcal{W} &= Sigmoid(Conv(F^{0}_{0 \rightarrow t},F^{0}_{1 \rightarrow t})),\\
      F_{t} &= \mathcal{W}\otimes F^{0}_{0 \rightarrow t}+(1-\mathcal{W})\otimes  F^{0}_{1 \rightarrow t}.
    \end{aligned}
    \label{f12}
\end{equation}
We note that this muti-scale forward warping is similar to SoftSplat\cite{2020Softmax} in form. However, our method differs in three aspects. First,  SoftSplat employs a nested gridnet\cite{2018Context} to fill the holes invoked by forward warping. Such a dense connection module would be an acceptable complexity if only two adjacent frames were used to interpolate the middle frame. But our model may accept many frames as input while also interpolating multiple frames between every adjacent two frames, in which case gridnet would be too time-consuming and memory intensive. To this end, we directly use the two warped features to complement each other, which only involves several convolutions for feature blending. Second, SoftSplat utilizes PWCnet\cite{2017PWC} as the optical flow estimator, which is also employed by many VFI methods. Still, we use a very light flow estimator Spynet\cite{ranjan2017optical}, which is seldom chosen by flow-based VFI methods. As a result, the accuracy of optical flow estimation is relatively poor. But the accurate estimation of optical flow is crucial for the VFI task. We still choose to use Spynet because 1) the optical flows are already calculated in stage 1 and do not need to be repeatedly estimated. 2) We do not want to use too complicated optical flow modules that would affect the model's efficiency. 3) to compensate for possible inaccurate estimates of optical flow, deformable convolution is further introduced to approximate finer details.
Till now, the temporal modulation process has been completed, and the features of pre-existing frames (in stage 1) and temporally interpolated frames(in stage 2) are ready to be upsampled in the next stage.
  \subsection{  Scale-arbitrary Upsampling }\label{upsample}
 \begin{figure}[t]
    \centering
    \includegraphics[width=9cm]{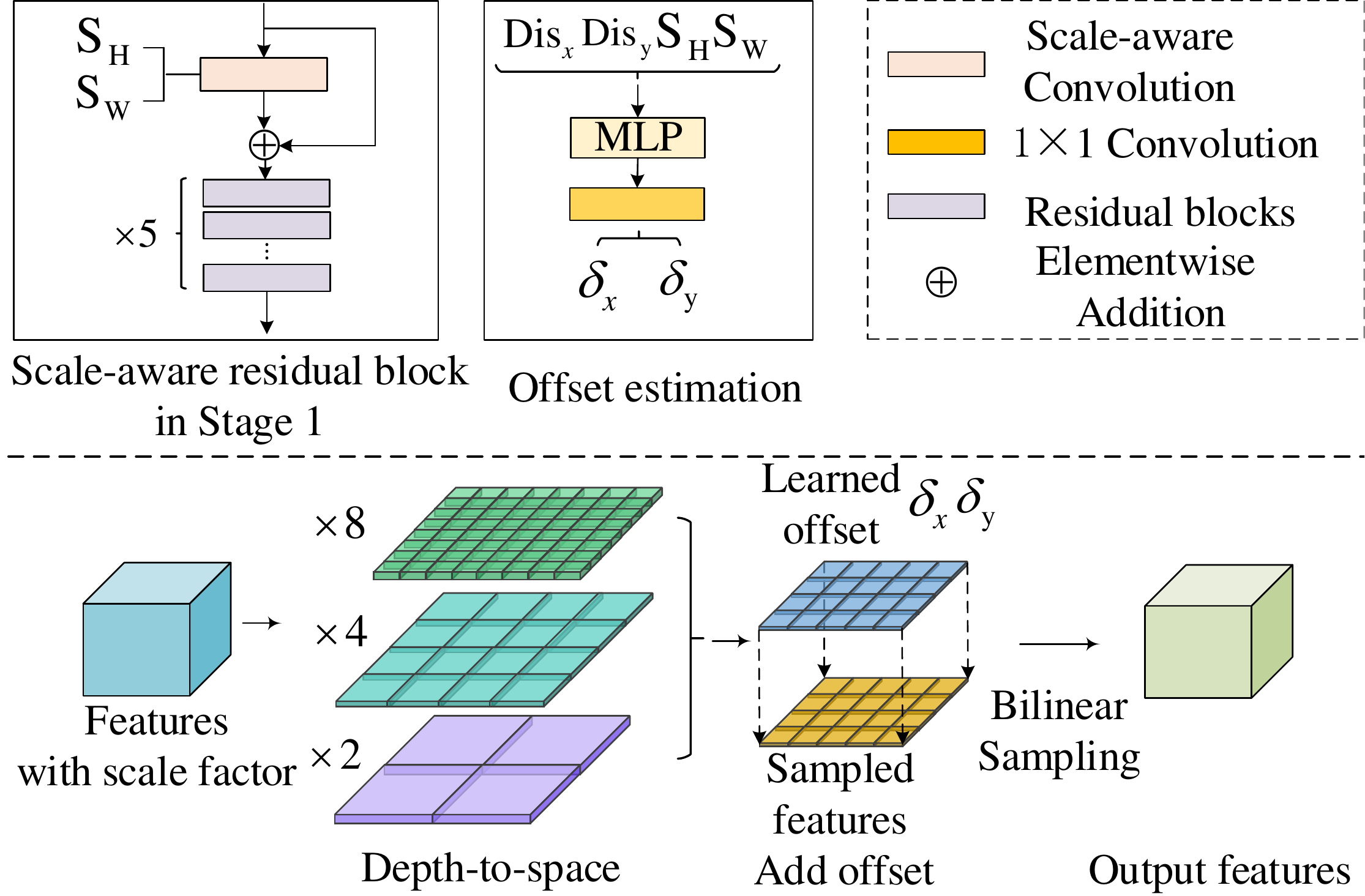}
    \caption{ Muti-scale feature blending module. We first perform a depth-to-space transformation to obtain features that expand to integral multiples of their original size. Afterward, we feed the relative position offsets of the pixels into a multilayer perceptron, allowing the model to adapt to different expanding scales. The learned offset and features are then combined and interpolated to the target space size via a bilinear interpolation function to get the output features.
    } \label{fig:3}
\end{figure}
Scale-arbitrary upsamples for single image super-resolution(SISR) have been explored by much previous work\cite{hu2019meta,wang2021learning,chen2021learning}. But in contrast, there is much less exploration related to scale-arbitrary video super-resolution. The straightforward idea is that video super-resolution does not need to be explicitly studied. Just taking any off-the-shelf magnification-arbitrary upsampling module can fulfill this goal. But this is not the case because most of these methods require estimating a unique sampling weight for each position. Such estimation is so memory intensive that it is almost impossible to process video with multiple input frames using an ordinary graphics card(e.g., NVIDIA 1080Ti GPU). In Meta-SR\cite{hu2019meta}, to achieve scale-arbitrary upsampling, Hu \emph{et al.} propose a meta-upscale module that takes a sequence of coordinate-related and scale-related vectors as input to predict the convolutional weights for every pixel point. Based on Meta-SR, Wang \emph{et al.} \cite{wang2021learning} further propose a plug-in module to generate dynamic scale-aware filters, which enable the network to adapt to an arbitrary scale factor. Although their methods achieve satisfactory results at different scales, their memory consumption is still unacceptable for our task. The memory footprint of spatially-varying filtering\cite{wang2021learning}  can be very high. (For a 720P HR image, if we set kernel size $k$ as 3, it will cost about 31.6G memory, even if we set $k=1$, it still cost about 3.5G)  Based on the above discussion, our motivation is to design a memory-saving and efficient upsampling module.

Several candidate operations exist to realize upsampling, including 1)  bilinear upsampling, 2) deconvolution with stride 3) pixel shuffle operation. The bilinear re-scaling method
is more flexible because it can be applied to an arbitrary scale
ratios. However, applying bilinear upsampling for a super-resolved RGB image may lead to severe performance degradation. Among the learning-based methods, we choose to adopt Pixelshuffle\cite{2016Real} as our basic module since it only involves the rearrangement of elements of input features and does not introduce additional learnable parameters. Almost all recent video super-resolution methods have adopted this approach. However, this depth-to-space operation does not accommodate arbitrary sizes. To this end, we propose a cascaded depth-to-space module, which allows the network to learn LR to HR mappings at different scales in an memory-efficient manner while maintaining scale flexibility. As shown in Fig.~\ref{fig:3}, in stage 1, we first employ multiple scale-aware residual blocks to extract features. Every block contains a scale-aware convolution\cite{wang2021learning} followed by several residual convolutions. In this way, features of different depths are integrated with scale information. In stage 3, we first perform the subpixel shuffling operator in three different scales:
  \begin{equation}
    \begin{aligned}
       Feat_{ps}^{l} &=  \mathcal{PS}(Fea\_in,{2}^{l}) , \quad l \in \{1,2,3\}
    \end{aligned}
    \label{f13}
\end{equation}
where $\mathcal{PS}$ means an periodic shuffling operator\cite{2016Real} that rearranges the elements of $H \times W \times C \cdot r^{2}$
tensor to a tensor of shape $r \cdot H \times r \cdot W \times C$. Note that we have transformed the feature's channel dimension $C$ of each scale before the $\mathcal{PS}$ operation to ensure that the results obtained at different scales have the same number of channels. (Here, we set $C=32$.)
 \begin{figure*}[htbp]
	\centering
	\includegraphics[width=16cm]{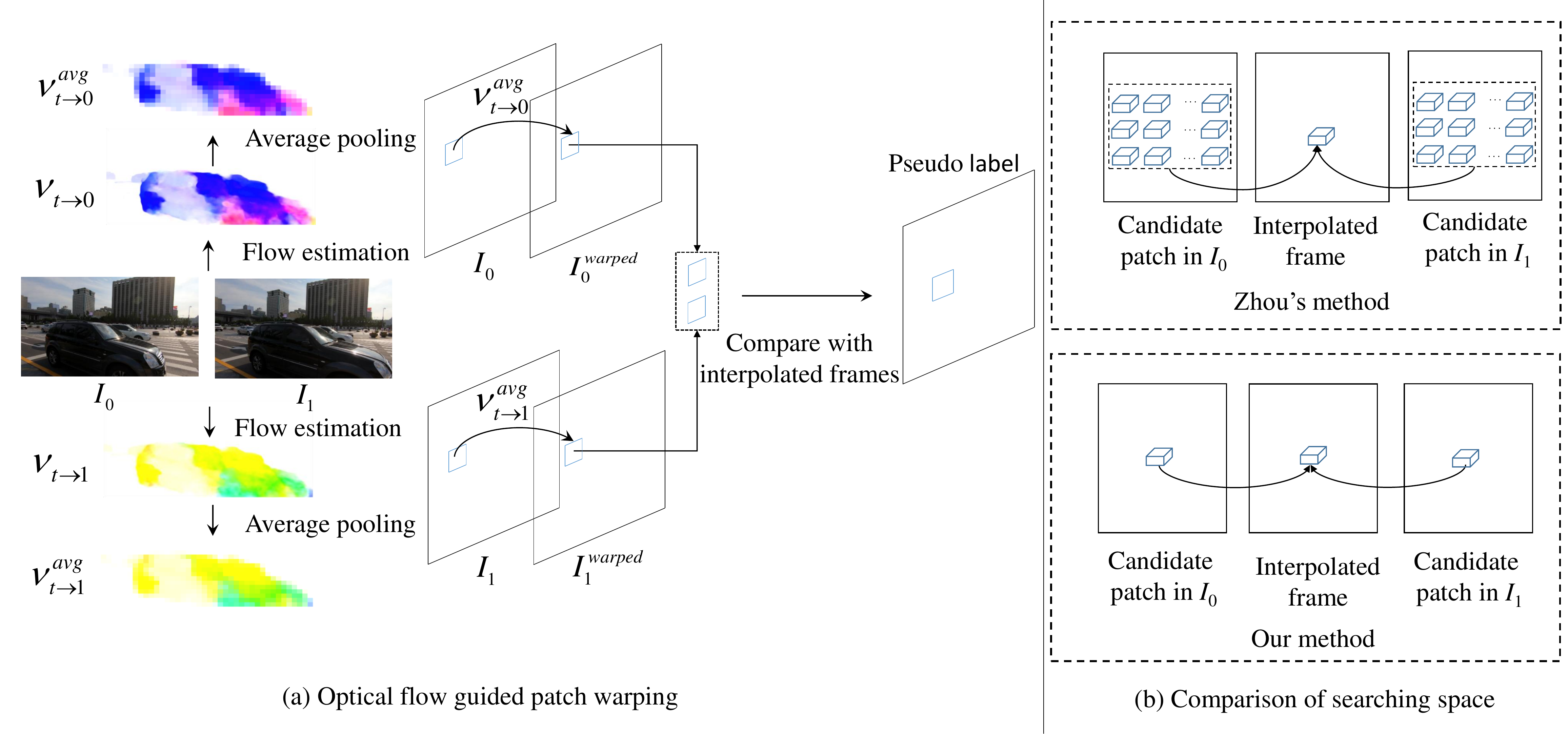}
	\caption{(a)  Schematic for optical flow guided patch warping, which aims to generate the pseudo label for texture consistency loss. (b) Comparison of patch searching space. Zhou's method\cite{zhou2022exploring} needs to search candidate patches from two reference frames one by one. In contrast, our method can directly find desired patches via patch warping and dramatically reduces the search space.} \label{fig:4}
\end{figure*}

To implicitly enable the network to accommodate sampling at different scales, We follow \cite{wang2021learning} and compute the relative offset positions of input features and output a relative distance vector $[Dis_{x},Dis_{y}]$:
  \begin{equation}
    \begin{aligned}
      LR(\sigma) &= \frac{\sigma+0.5}{R_{\sigma}}-0.5,\\
      Dis_{\sigma} &= LR(\sigma)- \lfloor\frac{\sigma+0.5}{R_{\sigma}}\rfloor,
      \sigma \in \{x,y\}.
    \end{aligned}
    \label{f15}
\end{equation}
In this process, each pixel at location $(x,y)$ in HR space is mapped to the LR space to estimate its coordinates($LR(x)$,$LR(y)$) and relative distance($Dis_{x}$,$Dis_{x}$), where $\lfloor \dots \rfloor$ means the rounded down operator. Relative distance vectors ($Dis_{x}$,$Dis_{x}$) and scale factors ($S_{x}$,$S_{y}$) are fed into a small MLP to get the offset map $\delta_{x}$,$\delta_{y}$. Next, we add the learned offset to the $Feat_{ps}$ to get the scale-adaptive features $Feat_{adapt}$. Then, a bilinear interpolation function is adopted to modulate the features to the target scale:
\begin{equation}
	\begin{aligned}
		Feat^{l} &=  bilinear( Feat_{adapt}^{l}, \frac{S_{H}}{{2}^{l}} ,\frac{S_{W}}{{2}^{l}}), \quad l \in \{1,2,3\},
	\end{aligned}
	\label{f14}
\end{equation}
where $S_{H}$ and $S_{W}$ denote the height and width scaling factor. Finally, we utilize several convolution layers for blending these features and getting the super-resolved color images. 
It is worth emphasizing that this cascading depth-to-space module is very conceptually straightforward. Since we do not intend to design complex and elaborate conditional spatial resampling modules for each pixel, nor do we want to argue about which upsampling module is the best. Here we just want to design a memory-efficient and practical module for scale-arbitrary upsampling that can achieve a small memory footprint and handle multiple input frames. Because we find that long-range temporal information is essential for ST-VSR, please refer to the experiment section for more information.
 \subsection{ Flow-guided Context-consistency Loss } \label{FGloss}
 Since we do not apply complex post-processing or context-refilling modules for temporal feature interpolation, it was still challenging to achieve satisfactory results when dealing with extreme motions and complex textures. A simple question arises  whether it is possible to further improve the quality of intermediate frames without increasing the inference computation burden. To address this problem,  Zhou \emph{et al.} \cite{zhou2022exploring} propose a texture consistency loss to generate a pseudo intermediate frame to allow the diversity of interpolated content. Their core idea is that existing deep VFI approaches strongly rely on the ground truth of intermediate frames, which sometimes ignore the non-unique motions in the natural world. So, they aim to generate a patch-level pseudo-label for the intermediate frame, which contains the most similar patch of the interpolated intermediate frame and the reference frames from both sides. It is conceptually similar to the frame inter-prediction technique in video coding(Advanced Motion Vector Prediction in HEVC\cite{HEVC}). 
 
But directly borrowing the texture consistency loss is, in fact, infeasible in our task. As mentioned before, since our model accepts multiple frames as input and interpolates multiple intermediate features between two adjacent frames, such a large number of patch searching will result in a huge computation burden. To solve this problem, we propose an optical flow-guided texture consistency loss. In reality, the optical flow estimation itself is a search process for similar regions between two frames. Meanwhile, the optical flow between the two adjacent frames has already been generated in Stage 1, so we decide to use optical flow as a priori knowledge to search for the most similar patches more efficiently. specifically, given the estimated motion fields $\mathcal{V}_{0 \rightarrow 1}$ and $\mathcal{V}_{1 \rightarrow 0}$ for two frames. We utilize the  complementary flow reversal(CFR) layer in XVFI\cite{sim2021xvfi} to approximate the intermediate motion flow $\mathcal{V}_{t \rightarrow 0}$ and $\mathcal{V}_{t \rightarrow 1}$:
\begin{equation}
    \begin{aligned}
          \mathcal{V}_{t \rightarrow 0} , \mathcal{V}_{t \rightarrow 1} =  CFR(\mathcal{V}_{0 \rightarrow 1},\mathcal{V}_{1 \rightarrow 0})\\
    \end{aligned}
    \label{f5}
\end{equation}

After obtaining the intermediate optical flow $\mathcal{V}_{t \rightarrow 0}$ and $\mathcal{V}_{t \rightarrow 1}$, the next step is to use them to warp the two adjacent frames' most similar patches to the intermediate frame. Note that the warping here is not pixel-wise alignment in the usual sense but patch-level alignment. The proposed optical flow-guided patch warping is illustrated in Fig.~\ref{fig:4}. Specifically, we first perform a pooling operation on the optical flow to get the average movement within a patch:
\begin{equation}
    \begin{aligned}
        \mathcal{V}^{avg}_{t \rightarrow 0} &= AvgPool(\mathcal{V}_{t \rightarrow 0},patch\_size = p), \\
        \mathcal{V}^{avg}_{t \rightarrow 1} &= AvgPool(\mathcal{V}_{t \rightarrow 1},patch\_size = p),
    \end{aligned}
    \label{f6}
\end{equation}
where $AvgPool$ denotes the average pooling operation and the patch size is set by default to 4 here.
Then, the reference frames from both sides are directly warped to the intermediate frame.
\begin{equation}
    \begin{aligned}
        I_{0 \rightarrow t}^{warped} &=  \mathcal{BW}(I_{0},\mathcal{V}^{avg}_{t \rightarrow 0}),  \\
        I_{1 \rightarrow t}^{warped} &= \mathcal{BW}(I_{1},\mathcal{V}^{avg}_{t \rightarrow 1})  
    \end{aligned},
    \label{f7}
\end{equation}
where $\mathcal{BW}$ denotes backward warping. It is worth noticing that as we perform patch-level rather than pixel-level warping, the optical flow does not need to be particularly accurate.
Finally, the two warped frames are compared patch by patch with the  intermediate frame and the most similar patch is selected as the pseudo label. We assume there are $N$ candidate patches to be matched, then, the matching process can be formulated as:
\begin{equation}
    \begin{aligned}
      k^{*},i^{*} &= \underset{{k \in (1,N),i \in \{0,1\}}} { \arg \min  } \quad
  \mathcal{L}2(P_{k}^{i},P_{k}^{pred}) \\
 \quad I_{t}^{pseudo} &= \bigcup\limits_{k^{*} = 1}^{ N } P_{k^{*}}^{i^{*}}
    \end{aligned}
    \label{f8}
\end{equation}
\begin{table*}[t]
	\caption{Quantitative comparisons of PSNR (dB), SSIM, speed (FPS), on Vid4, REDS and Vimeo-90K-T. The inference time is calculated on Vid4 dataset with one Nvidia 1080Ti GPU. The best two results are highlighted in {\color{red}red} and {\color{blue}blue} colors.}
	\label{tab:1}
	\centering
	\resizebox{0.98\linewidth}{!}{
		\begin{tabular}{c|cc|cc|cc|cc|cc|cc|c|c}
			\hline
			VFI+(V)SR/ST-VSR methods & \multicolumn{2}{c|}{\begin{tabular}[c]{@{}c@{}}Vid4\\ PSNR \quad SSIM\end{tabular}} & \multicolumn{2}{c|}{\begin{tabular}[c]{@{}c@{}}REDS\\ PSNR \quad SSIM\end{tabular}} & \multicolumn{2}{c|}{\begin{tabular}[c]{@{}c@{}}Vimeo-Fast\\ PSNR \quad SSIM\end{tabular}} & \multicolumn{2}{c|}{\begin{tabular}[c]{@{}c@{}}Vimeo-Medium\\ PSNR \quad SSIM\end{tabular}} & \multicolumn{2}{c|}{\begin{tabular}[c]{@{}c@{}}Vimeo-Slow\\ PSNR \quad SSIM\end{tabular}} & \multicolumn{2}{c|}{\begin{tabular}[c]{@{}c@{}}Vimeo-Total\\ PSNR \quad SSIM\end{tabular}} & \begin{tabular}[c]{@{}c@{}}Speed\\ FPS\end{tabular} & \begin{tabular}[c]{@{}c@{}}Parameters\\ millions\end{tabular} \\ \hline \hline
			SuperSloMo\cite{0Super} +Bicubic      & 22.84                                & 0.5772                                 & 25.23                                & 0.6761                                 & 31.88                                   & 0.8793                                    & 29.94                                    & 0.8477                                     & 28.73                                   & 0.8102                                    & 29.99                                    & 0.8449                                    & -                                                   & 19.8                                                          \\
			SuperSloMo\cite{0Super} +RCAN\cite{2018Image}         & 23.78                                & 0.6397                                 & 26.37                                & 0.7209                                 & 34.52                                   & 0.9076                                    & 32.50                                    & 0.8844                                     & 30.69                                   & 0.8624                                    & 32.44                                    & 0.8835                                    & 1.91                                                & 19.8+16.0                                                     \\
			SuperSloMo\cite{0Super}+RBPN\cite{2019Recurrent}          & 23.76                                & 0.6362                                 & 26.48                                & 0.7281                                 & 34.73                                   & 0.9108                                    & 32.79                                    & 0.8930                                     & 30.48                                   & 0.8354                                    & 32.62                                    & 0.8839                                    & 1.55                                                & 19.8+12.7                                                     \\
			SuperSloMo\cite{0Super} +EDVR\cite{2019EDVR}         & 24.40                                & 0.6706                                 & 26.26                                & 0.7222                                 & 35.05                                   & 0.9136                                    & 33.85                                    & 0.8967                                     & 30.99                                   & 0.8673                                    & 33.45                                    & 0.8933                                    & 4.94                                                & 19.8+20.7                                                     \\ \hline
			Sepconv\cite{2017Video}+Bicubic          & 23.51                                & 0.6273                                 & 25.17                                & 0.6760                                 & 32.27                                   & 0.8890                                    & 30.61                                    & 0.8633                                     & 29.04                                   & 0.8290                                    & 30.55                                    & 0.8602                                    & -                                                   & 21.7                                                          \\
			Sepconv\cite{2017Video} +RCAN\cite{2018Image}            & 24.92                                & 0.7236                                 & 26.21                                & 0.7177                                 & 34.97                                   & 0.9195                                    & 33.59                                    & 0.9125                                     & 32.13                                   & 0.8967                                    & 33.50                                    & 0.9103                                    & 1.86                                                & 21.7+16.0                                                     \\
			Sepconv\cite{2017Video} +RBPN\cite{2019Recurrent}             & 26.08                                & 0.7751                                 & 26.32                                & 0.7254                                 & 35.07                                   & 0.9238                                    & 34.09                                    & 0.9229                                     & 32.77                                   & 0.9090                                    & 33.97                                    & 0.9202                                    & 1.51                                                & 21.7+12.7                                                     \\
			Sepconv\cite{2017Video} +EDVR\cite{2019EDVR}            & 25.93                                & 0.7792                                 & 26.14                                & 0.7205                                 & 35.23                                   & 0.9252                                    & 34.22                                    & 0.9240                                     & 32.96                                   & 0.9112                                    & 34.12                                    & 0.9215                                    & 4.96                                                & 21.7+20.7                                                     \\ \hline
			DAIN\cite{DAIN} +Bicubic            & 23.55                                & 0.6268                                 & 25.22                                & 0.6783                                 & 32.41                                   & 0.8910                                    & 30.67                                    & 0.8636                                     & 29.06                                   & 0.8289                                    & 30.61                                    & 0.8607                                    & -                                                   & 24                                                            \\
			DAIN\cite{DAIN} +RCAN\cite{2018Image}               & 25.03                                & 0.7261                                 & 26.33                                & 0.7233                                 & 35.27                                   & 0.9242                                    & 33.82                                    & 0.9146                                     & 32.26                                   & 0.8974                                    & 33.73                                    & 0.9126                                    & 1.84                                                & 24.0+16.0                                                     \\
			DAIN\cite{DAIN} +RBPN\cite{2019Recurrent}                & 25.96                                & 0.7784                                 & 26.57                                & 0.7344                                 & 35.55                                   & 0.9300                                    & 34.45                                    & 0.9262                                     & 32.92                                   & 0.9097                                    & 34.31                                    & 0.9234                                    & 1.43                                                & 24.0+12.7                                                     \\
			DAIN\cite{DAIN} +EDVR\cite{2019EDVR}               & 26.12                                & 0.7836                                 & 26.39                                & 0.7291                                 & 35.81                                   & 0.9323                                    & 34.66                                    & 0.9281                                     & 33.11                                   & 0.9119                                    & 34.52                                    & 0.9254                                    & 4                                                   & 24.0+20.7                                                     \\  \hline
			STARnet\cite{2020Space}                  & 26.06                                & 0.8046                                 & 26.39                                & 0.7444                                 & 36.19                                   & 0.9368                                    & 34.86                                    & 0.9356                                     & 33.10                                   & 0.9164                                    & 34.71                                    & 0.9318                                    & 10.54                                               & 111.61                                                        \\
			Zooming Slow-Mo\cite{2021Zooming}          & 26.31                                & 0.7976                                 & 26.72                                & 0.7453                                 & 36.81                                   & 0.9415                                    & 35.41                                    & 0.9361                                     & 33.36                                   & 0.9138                                    & 35.21                                    & 0.9323                                    & \textcolor{blue}{12.4}                                                & \textcolor{red}{11.1}                                                          \\
			TMnet\cite{xu2021temporal}                    & 26.43                                & 0.8016                                 & 26.81                                & 0.7476                                 & \textcolor{blue}{37.04}                                   & 0.9435                                    & 35.60                                    & 0.9380                                     & 33.51                                  & 0.9159                                    & 35.39                                    & 0.9343                                    & 11.6                                                & \textcolor{blue}{12.26}                                                         \\ 
			EBFW\cite{zhangVCIP} (Our previous work)                    & \textcolor{blue}{26.74}                                & \textcolor{blue}{0.8175}                                 & \textcolor{blue}{26.90}                               & \textcolor{blue}{0.7498}                                & 37.01                                   & \textcolor{blue}{0.9445}                                    & \textcolor{blue}{35.76}                                    & \textcolor{blue}{0.9400}                                     & \textcolor{blue}{33.63}                                 & \textcolor{blue}{0.9180}                                    & \textcolor{blue}{35.52}                                    & \textcolor{blue}{0.9362}                                    & \textcolor{red}{17.14 }                                               & 20.13                                                         \\ \hline
			C-STVSR-fix (Ours)           & \textcolor{red}{26.87}                                & \textcolor{red}{0.8213}                                 & \textcolor{red}{26.99}                                & \textcolor{red}{0.7525}                                 & \textcolor{red}{37.09}                                   & \textcolor{red}{0.9447}                                    & \textcolor{red}{35.82}                                    & \textcolor{red}{0.9405}                                     & \textcolor{red}{33.69}                                   & \textcolor{red}{0.9184}                                    & \textcolor{red}{35.57}                                    &\textcolor{red}{0.9365}                                    & 8.03                                              & 13.19                                                         \\ \hline
	\end{tabular}}
\end{table*}
\begin{figure*}[t]
	\centering
	\includegraphics[width=15cm]{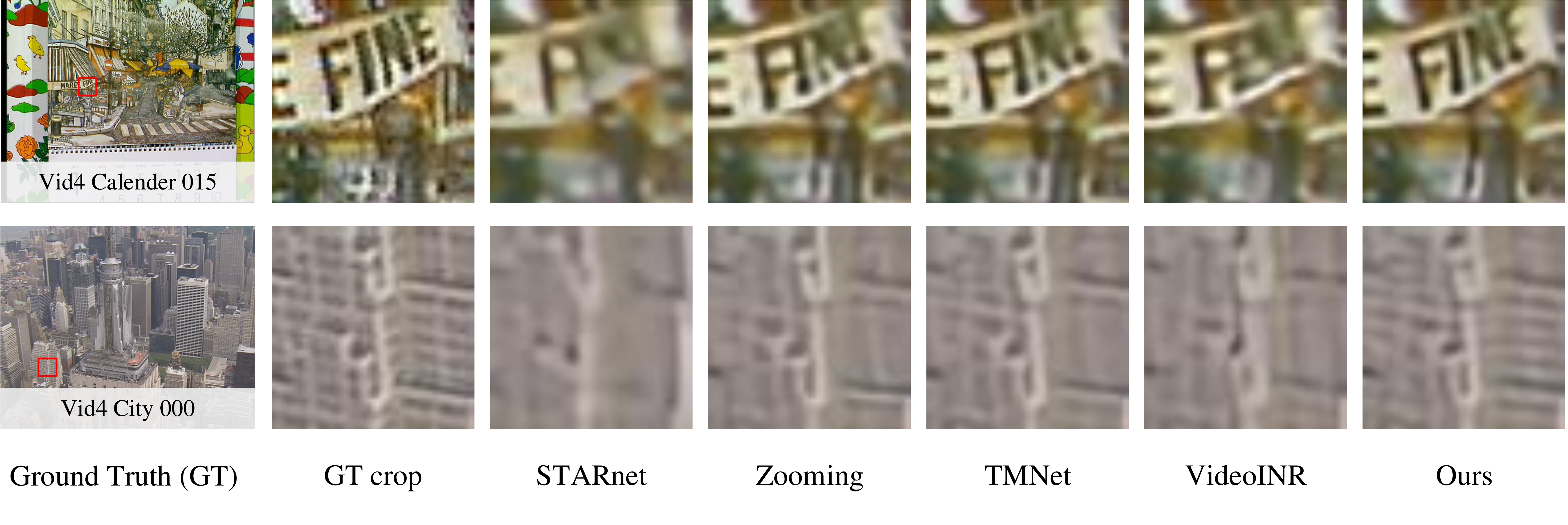}
	\caption{Visual results on Vid4 of different methods for space 4$\times$ and time 2$\times$ super resolution. Parts of areas are zoomed in to facillate comparison.} \label{fig:5}
\end{figure*}
where $P_{k}^{i}$ is the $k$-th patch from the $i$-th reference frames after census transform\cite{1997Non},  and the final pseudo-label $I_{t}^{pseudo}$ is the concatenation of the patches from the two reference images. For each patch in the pseudo label, the candidate searching number is reduced from 2*$N$ to 2. It is worth emphasizing that the optical flow has already been estimated in stage one and does not need to be estimated repeatedly. So the pseudo-label generation process is very efficient.
In addition, the operations involved in the patch warping process, such as image partition and optical flow pooling, are easy to implement with modern deep learning frameworks, and the whole process is performed parallelly.

\section{Experiments}

\subsection{ Experimental Setup}

\noindent \textbf{Dataset}
We adopt  Vimeo-90K-T\cite{DBLP:journals/corr/abs-1711-09078} to train our model. Vimeo-90k-T contains 91701 video clips, each consisting of seven consecutive frames at the resolution of 448 $\times$ 256. 
We follow \cite{2021Zooming,xu2021temporal} and split the Vimeo-90K-T  test set into three subsets of the fast, medium, and slow motion according to their average motion magnitude. Some all-black sequences are removed from the test set since they will lead to infinite values on PSNR.
We use various datasets including Vimeo-90K-T\cite{DBLP:journals/corr/abs-1711-09078}, Vid4\cite{5995614}, REDS\cite{Nah_2019_CVPR_Workshops}, Adobe240FPS\cite{2019Video}, SPMCS\cite{tao2017detail}  for evaluation. To be specific, we first conduct fixed spatial-temporal super-resolution experiments on Vimeo-90K-T, Vid4, and REDS. To further validate the model's ability to handle large motions and synthesize continuous temporal trajectories, we perform experiments on two challenging datasets: REDS and  Adobe240FPS. Finally, we compare the performance of continuous space-time super-resolution on the SPMCS dataset. The detailed experimental setup is described in the corresponding section. \\
%
%
%
%
\noindent \textbf{Implemention Details}
We utilize the Adam \cite{kingma2014adam} with $\beta_{1}$ = 0.9 and $\beta_{2}$ = 0.999 as optimizer. The learning rate is initialized as
2 $\times$ 10$^{-4}$ and is decayed to 1 $\times$ 10$^{-7}$ with a cosine annealing scheduler. When training the model, we downsample the odd-index HR frames with bicubic interpolation to obtain the LR frames as the inputs. To allow longer
propagation, we follow the setup of\cite{2020BasicVSR} and perform temporal
augment by flipping the original input sequence. For a fair comparison, we adopt different strategies and train two models, which we refer to as model-fix and model-continuous. The model-fix is trained with a fixed downsample factor of 4. When training model-continuous, we randomly sample scales in distribution $\{2.0,2.2 \dots 3.8,4.0\}$ with a stride of 0.2.
 To optimize the model, we use different loss functions to supervise pre-existing and temporally interpolated frames. For the former, we select the Charbonnier loss function\cite{lai2017deep}, as suggested in \cite{2021Zooming,xu2021temporal}:
   \begin{equation}
    \begin{aligned}
        loss_{exist} = \sqrt{{\Vert I^{exist}-I^{GT} \Vert}^{2}+{\epsilon}^{2}}.
    \end{aligned}
    \label{f16}
\end{equation}
  For the latter, we relax the precise correspondence between the intermediate synthesis frames and the ground truth (GT). So the optimization goal is not only $\mathcal{L}_{1}$ between GT and the prediction frames but also considers the same continuous semantic pattern of synthesis and reference frames from both sides. Our total learning goal is formulated as follows:
   \begin{equation}
    \begin{aligned}
        loss_{inter} =\mathcal{L}_{1}(I^{pred},I^{GT})+ \alpha \mathcal{L}_{1}(I^{pred},I^{pseudo}),
    \end{aligned}
    \label{f17}
\end{equation}
where $I^{pred}$ denotes the predicted intermediate frame, and $I^{GT}$ and $I^{pseudo}$ refer to the ground truth and the pseudo label. $\alpha$ is a hyperparameter to balance the 
importance of the two items. We follow \cite{zhou2022exploring} and set it to 0.1. So the total loss function can be described as:
   \begin{equation}
    \begin{aligned}
        loss_{total} =loss_{exist}+loss_{inter} 
    \end{aligned}
    \label{f18}
\end{equation}

We use Peak Signal-to-Noise Ratio (PSNR) and Structural Similarity Index (SSIM) for evaluation. The inference speed of these methods is measured on the entire Vid4  dataset using one Nvidia 1080Ti GPU.
\subsection{ Comparison with State-of-the-arts methods}

\begin{figure*}[t]
	
	\centering
	\includegraphics[width=15cm]{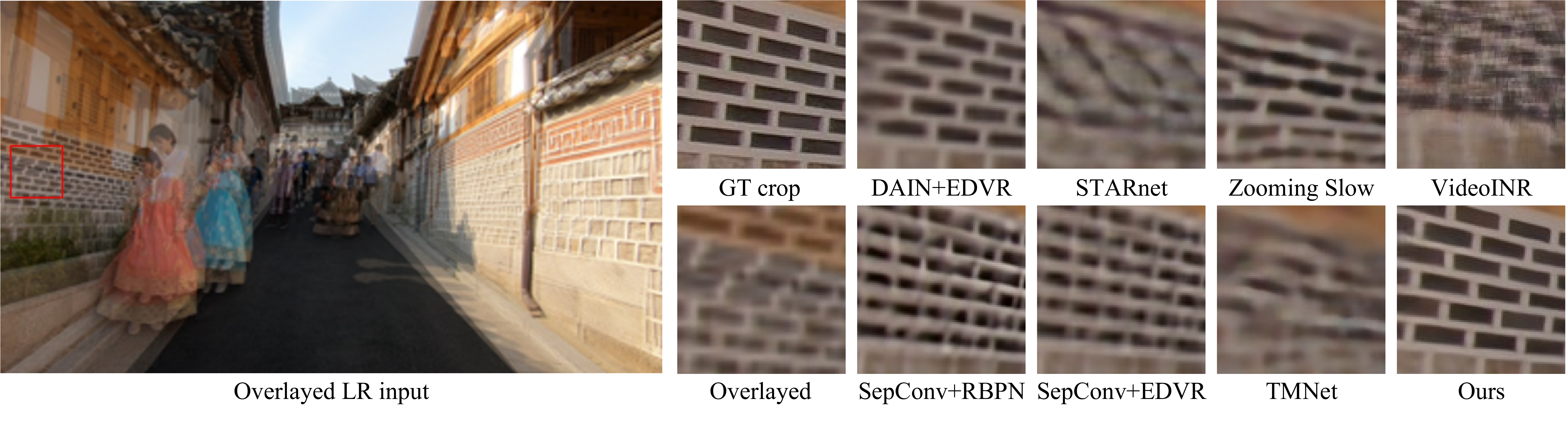}
	\caption{  Visual comparison of sythetic intermediate frames under extreme motions on REDS testset. Our method can synthesis more visually plausible results  } \label{fig:6}
\end{figure*}

\begin{figure*}[t]
	\centering
	\includegraphics[width=16cm]{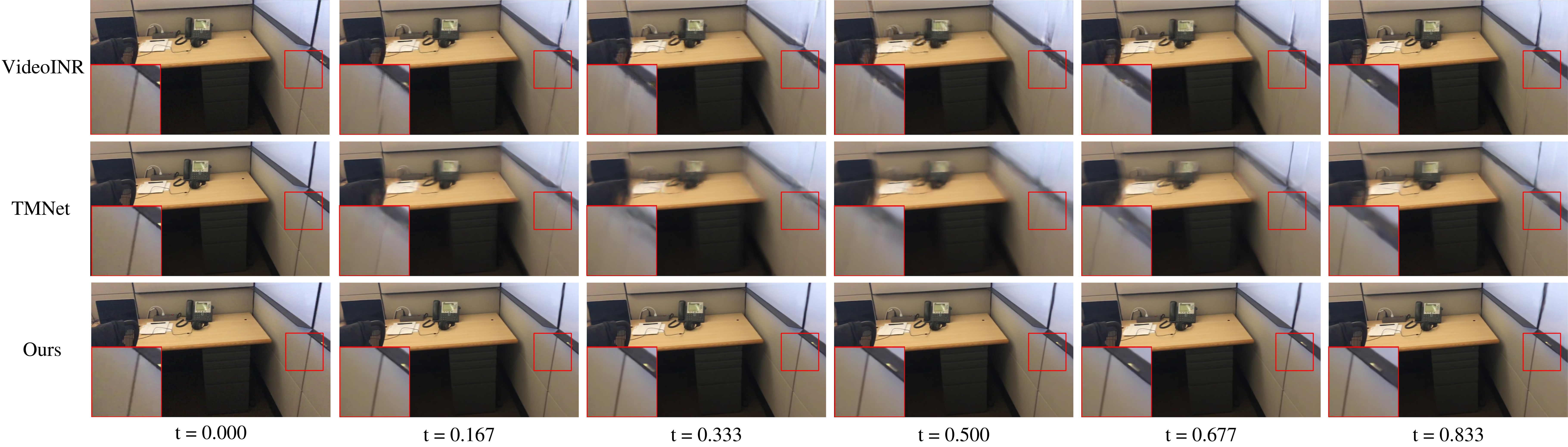}
	\caption{ Visual comparison of temporal consistency. We display the reconstruction results of pre-existing frames and five temporal interpolated frames here. Our method can synthesise continuous and consistent trajectories.   } 	\label{fig:7}
\end{figure*}

\begin{figure*}
	\centering
	\includegraphics[width=18cm]{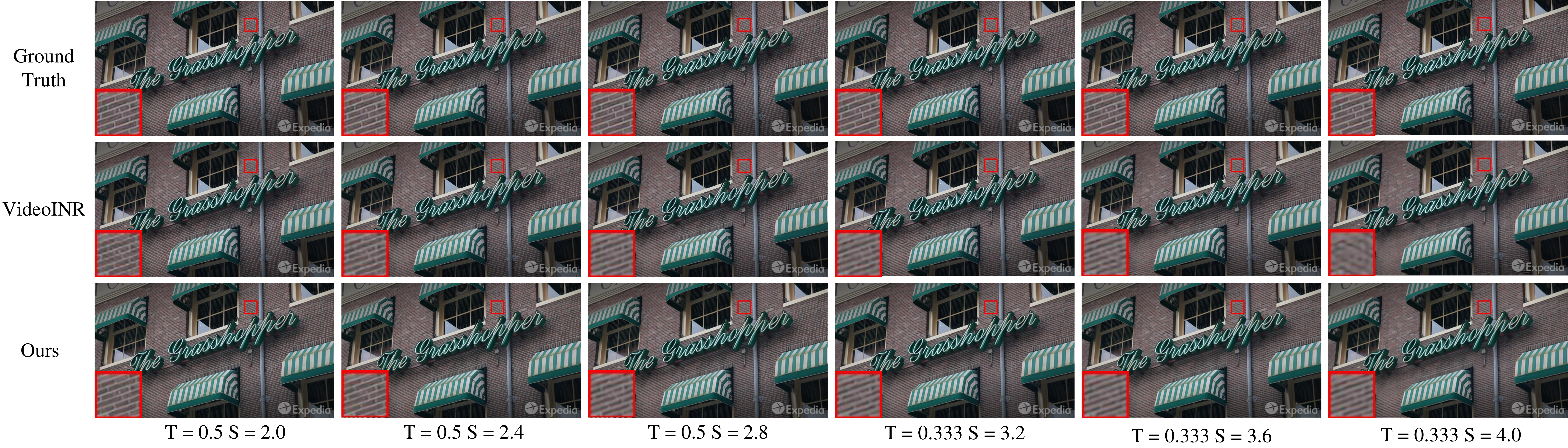}
	\caption{ Qualitative comparison with VideoINR for continuous space-time video super-resolution on SPMCS.  T=A , S=B refers to A temporal intermediate time and B  up-sampling  scale. } \label{fig:arb}
\end{figure*}

\noindent \textbf{Comparison with fixed spatio-temporal upsampling methods.}
Since the vast majority of current methods can only accomplish space-time upsampling with fixed spatial and temporal interpolation scales, we first need to compare our model with these methods. Specifically, we compare the  results for space 4$\times$  and time 2$\times$  super-resolution with existing state-of-the-art one-stage approaches\cite{2020Space,2021Zooming,xu2021temporal} and two-stage approaches which sequentially apply separated VFI\cite{0Super,2017Video,DAIN} and (V)SR\cite{2018Image,2019Recurrent,2019EDVR}. All these methods are tested on three datasets:Vid4\cite{5995614}, Vimeo-90k-T\cite{DBLP:journals/corr/abs-1711-09078}, and REDS\cite{Nah_2019_CVPR_Workshops}. which are commonly used for VSR and VFI. For a fair comparison, we remove the scale-aware convolution in the feature extraction module in stage 1 and fix the spatial up-sampling factor when training our model. From Table~\ref{tab:1}, one can see that the proposed method outperforms all the other
\begin{table*}[t]
	\centering
	\caption{  Quantitative assessment(PSNR (dB) / SSIM) of multiple-frame interpolation  of different methods on Adobe240FPS testset.  The top results are highlighted with {\color{red} red} color.
	}
	\resizebox{0.93\linewidth}{!}{
		\begin{tabular}{c|c|c|c|c|c|c}
			\hline
			Time stamp     & 0.000 & 0.167 & 0.333 & 0.500 & 0.667 & 0.833 \\ \hline \hline
			TMNet\cite{xu2021temporal}         &31.50 / 0.8924 & 27.00 / 0.8101 & 25.47 / 0.7691 & 25.44 / 0.7669 & 25.50 / 0.7703 & 26.86 / 0.8062 \\ \hline 
			VideoINR\cite{chen2022videoinr}      &30.75 / 0.8889 & 27.58 / 0.8301 & 26.17 / 0.7963 & 25.96 / 0.7890 & 26.19 / 0.7953 & 27.53 / 0.8283 \\ \hline
			EBFW\cite{zhangVCIP}(our previous work) &32.89 / 0.9195 & 27.99 / 0.8494 & 27.61 / 0.8346 & 27.68 / 0.8354 & 27.67 / 0.8367 & 28.05 / 0.8507 \\ \hline
			C-STVSR (Ours)           &\textcolor{red}{33.15} / \textcolor{red}{0.9236} & \textcolor{red}{28.13} / \textcolor{red}{0.8564} & \textcolor{red}{27.63} / \textcolor{red}{0.8374} & \textcolor{red}{27.74} / \textcolor{red}{0.8389} & \textcolor{red}{27.72} / \textcolor{red}{0.8403} & \textcolor{red}{28.29} / \textcolor{red}{0.8591} \\ \hline
		\end{tabular}
		\label{tab:2}}
\end{table*}
methods on all quantitative evaluation indicators and has similar runtime and model size. Compared to our previous work\cite{zhangVCIP}, the new model saves about 35\%  the number of parameters and achieves higher performance. However, the multi-scale deformable alignments also somewhat affect the inference speed. Notably,  it outperforms  TMnet\cite{xu2021temporal} by 0.44 dB in PSNR on the Vid4\cite{5995614} dataset. The visual results of different methods are shown in Fig.~\ref{fig:5}. It can be seen that, in accordance with its significant quantitative improvements,  our model can reconstruct more visually pleasing
results with sharp edges and fine details than its competitors. 
\\
\textbf{Comparison for temporal consistency. }
We next compare the restoration results of different methods in temporal consistency. 
Unlike the Vanilla VSR task, the ST-VSR needs to restore two types of frames: the pre-existing and temporally interpolated frames. The former frames have their low-resolution image as the reference, so the restoration effect is more reliable. In contrast, the latter has no low-resolution reference, so the reconstruction results are relatively poor. This situation may result in quality fluctuations of the restoration video, especially when dealing with large motions. It is, therefore, essential to compare the visual quality of the interpolated intermediate frames. We first analyze the restoration results at $t$=0.5 on the REDS\cite{Nah_2019_CVPR_Workshops}. REDS is a very challenging dataset containing various scenarios and large motions. The objective results are given in Table~\ref{tab:1}, and one can see that our method outperforms all other methods in terms of PSNR and SSIM. We also compare the subjective results in Fig.~\ref{fig:6}. 
\begin{table*}[t]
	\caption{ Quantitative comparison (PSNR/SSIM) of continuous space-time super-resolution. The results are calculated on SPMCS.  T $\times$ A , S $\times$ B refers to A temporal interpolated frames  and B up-sampling scale.  The top results are highlighted with {\color{red} red} color.}\label{tab:3}
	\centering
	\resizebox{0.85\linewidth}{!}{\begin{tabular}{c|llllll}
			\hline
			\multicolumn{1}{l|}{} & T $\times$ 2 S $\times$ 2               & T $\times$ 2 S $\times$ 2.4             & T $\times$ 2 S$\times$ 2.8             & T $\times$ 2 S $\times$ 3.2             & T $\times$ 2 S $\times$ 3.6             & T $\times$ 2 S $\times$ 4.0             \\ \hline \hline
			VideoINR\cite{chen2022videoinr}              & \multicolumn{1}{c}{30.65 / 0.9024} & \multicolumn{1}{c}{29.86 / 0.9024} & \multicolumn{1}{c}{30.44 / 0.9001} & \multicolumn{1}{c}{29.52 / 0.8772} & \multicolumn{1}{c}{28.63 / 0.8622} & \multicolumn{1}{c}{28.61 / 0.8435} \\ \hline
			C-STVSR(Ours)                  & \multicolumn{1}{c}{\textcolor{red}{35.94}  / \textcolor{red}{0.9672}}                                 & \multicolumn{1}{c}{\textcolor{red}{34.33} / \textcolor{red}{0.9504}}                                  &\multicolumn{1}{c}{\textcolor{red}{33.02} / \textcolor{red}{0.9308}}                                   & \multicolumn{1}{c}{\textcolor{red}{31.98} / \textcolor{red}{0.9093}}                                  & \multicolumn{1}{c}{\textcolor{red}{30.20} / \textcolor{red}{0.8831}}                                  & \multicolumn{1}{c}{\textcolor{red}{29.80} / \textcolor{red}{0.8600}}                                  \\ \hline
			& T $\times$ 4 S $\times$ 2               & T $\times$ 4 S $\times$ 2.4             & T $\times$ 4 S $\times$ 2.8             & T $\times$ 4 S $\times$ 3.2             & T $\times$ 4 S $\times$ 3.6             & T $\times$ 4 S $\times$ 4.0             \\ \hline
			VideoINR\cite{chen2022videoinr}              & \multicolumn{1}{c}{30.08 / 0.8907} & \multicolumn{1}{c}{29.49 / 0.8923} & \multicolumn{1}{c}{30.04 / 0.8902} & \multicolumn{1}{c}{29.27 / 0.8699} & \multicolumn{1}{c}{28.44 / 0.8551} & \multicolumn{1}{c}{28.44 / 0.8377}  \\ \hline
			C-STVSR(Ours)                   &  \multicolumn{1}{c}{\textcolor{red}{34.50} / \textcolor{red}{0.9561}}                                & \multicolumn{1}{c}{\textcolor{red}{33.16} / \textcolor{red}{0.9393}}                                 & \multicolumn{1}{c}{\textcolor{red}{31.96} / \textcolor{red}{0.9187}}                                 & \multicolumn{1}{c}{\textcolor{red}{31.12} / \textcolor{red}{0.8971}}                                 &\multicolumn{1}{c}{\textcolor{red}{29.51} / \textcolor{red}{0.8713}}                                   &       \multicolumn{1}{c}{\textcolor{red}{29.22} / \textcolor{red}{0.8484}}                           \\ \hline
	\end{tabular}}
\end{table*}
It can be seen that there exists extreme motion between two successive frames. Some  kernel-based methods\cite{2021Zooming,xu2021temporal,2017Video} suffers from severe distortion.
This may be because the offset estimated by the kernel-based method is limited by the receptive field of the convolution kernel. At the same time, the motion information is learned unsupervised, which is unreliable when estimating large motions. Some flow-based methods\cite{DAIN,chen2022videoinr} also fails to  restore sharp edges. In contrast, our method can recover the best details by benefiting from this coarse-to-fine motion estimation and motion compensation. More discussions and visual results are provided in the Ablation study section.  
Since our method can interpolate arbitrary intermediate frames, we then compare the performance of multi-frame interpolation. To verify its performance in temporal consistency, we compare the proposed method with TMNet\cite{xu2021temporal} and VideoINR\cite{chen2022videoinr}, both of which can also perform the modulated temporal interpolation.
Specifically, we follow TMnet and study space 4$\times$ and temporal 6$\times$ results on Adobe240FPS\cite{su2017deep} test set. Please note that TMnet and VideoINR require high frame rate training data in the training stage to supervise multiple intermediate moments and achieve the arbitrary intermediate interpolation. On the contrary, our method only supervises the middle moment $t$=0.5 in the training stage. As shown in Table~\ref{tab:2}, our method surpasses its competitors in objective metrics. In addition, our new model achieves higher performance with fewer parameters than our previous work(EBFW\cite{zhangVCIP}). We also provide a visual comparison in Fig.~\ref{fig:7}. As one can see, the screen moves rapidly from left to right,  TMNet and videoINR appear significantly blurred and fail to synthesize a continuous temporal trajectory. By contrast, our method can synthesize the most reliable and plausible results.
All these validation results can prove that our method has great advantages when approximating large motions and synthesizing continuous temporal trajectories.\\
\textbf{Comparison for continuous space-time super-resolution. }
We finally compare the results of continuous spatial-temporal upsampling.
As some previous work\cite{shi2021learning,chen2022videoinr} has pointed out, a two-stage method composed of successive video frame interpolation and image super-resolution performs significantly worse than a one-stage method. In addition, the continuous image super-resolution does not consider temporal information, so the comparison is not fair. Here, we only consider the comparison with the end-to-end spatio-temporal super-resolution method. As far as we know, VideoINR\cite{chen2022videoinr}   is the only work that can achieve continuous ST-VSR and release their code. So we choose to compare with it. However, VideoINR is not trained on Vimeo-90K-T\cite{DBLP:journals/corr/abs-1711-09078}, but on Adobe240FPS\cite{2019Video} since this method needs to be trained on a high-frame-rate dataset in order to achieve time-arbitrary modulation.  For a fair comparison, we followed its settings and retrained our model on the Adobe240FPS train set. The comparison results are provided in Table~\ref{tab:3}. One can see that our method outperform VideoINR at different combination of intermediate moments and scales. The visual perception also validates the objective results, as can be seen in Fig~\ref{fig:arb}. Our model can reconstruct more reliable results and finer details at different scales.
\begin{table*}[t]
	\caption{ Qualitative Comparison of (PSNR(dB))$\uparrow$ / average GPU memory usage (MB)$\downarrow$  on REDS testset when feeding different lengths of LR frames. The best performance is highlighted in \textcolor{red}{red}. Here we perform space 4$\times$ and time 2$\times$ interpolation. The memory usage is computed on the LR size of 180 $\times$ 320 using one Nvidia 1080Ti GPU. N/A (Not available) denotes  out-of-memory cases. }  
	\resizebox{1.00\linewidth}{!}{\begin{tabular}{c|c|c|c|c|c|c|c|c}
			\hline 
			\begin{tabular}[c]{@{}c@{}}Input\\ length\end{tabular} & 4            & 5            & 6            & 7                        & 10           & 17           & 26           & 51           \\ \hline \hline
			TMNet\cite{xu2021temporal}                                                  & 26.80 / 7590 & 26.82 / 7830 & 26.81 / 9692 & 26.81 / 10904             & N/A            & N/A            & N/A            & N/A            \\ \hline
			Zooming
			Slow-Mo\cite{2021Zooming}& 26.68 / 6096  & 26.72 / 7417 & 26.73 / 9154 &  26.72 / 10590 & N/A                        & N/A            & N/A            & N/A            \\ \hline
			C-STVSR (Ours)                                                    & \textcolor{red}{26.81} / \textcolor{red}{3091} & \textcolor{red}{26.83} / \textcolor{red}{3298} &  \textcolor{red}{26.85} / \textcolor{red}{3422} & \textcolor{red}{26.86} / \textcolor{red}{3660}   & \textcolor{red}{26.88}  / \textcolor{red}{4984} &\textcolor{red}{26.89} / \textcolor{red}{5462} & \textcolor{red}{26.90} / \textcolor{red}{6534} & \textcolor{red}{26.99} / \textcolor{red}{10965} \\ \hline
	\end{tabular}}
	\label{tab:length}
\end{table*}
\begin{table*}[t]
	\centering
	\caption{ Ablation on temporal modulation module. } \resizebox{0.80\linewidth}{!}{
		\begin{tabular}{c|c|c|c|c|c}
			\hline 
			& REDS           & Vimeo-Fast     & Vimeo-Medium   & Vimeo-Slow     & Vimeo-Total    \\ 
			& PSNR   SSIM & PSNR   SSIM & PSNR   SSIM & PSNR   SSIM & PSNR   SSIM \\ \hline  \hline
			w/o FWG & 26.65 / 0.7434 & 37.01 / 0.9429 & 35.57 / 0.9384 & 33.63 / 0.9180  & 35.40 / 0.9349 \\ \hline
			w/o DA & 26.85 / 0.7483 & 36.67 / 0.9410 & 35.60 / 0.9386 & 33.56 / 0.9169 & 35.35 / 0.9345 \\ \hline
			w/o FGL & 26.89 / 0.7494 & 36.80 / 0.9430 & 35.65 / 0.9396 & 33.58 / 0.9175 & 35.40 / 0.9356 \\ \hline
			C-STVSR(Ours)     & 26.99 / 0.7525 & 37.09 / 0.9447 & 35.81 / 0.9404 & 33.69 / 0.9183 & 35.57 / 0.9365 \\ \hline
	\end{tabular}}\label{tab:5}
\end{table*}
\subsection{Algorithmic Analyses}
We next analyze the proposed scheme's performance and efficiency and discuss our approach's advantages in mining long-range spatio-temporal dependencies.\\
\textbf{Comparison for long-range modeling ability. }
Much of previous work on VFI\cite{xu2019quadratic,kalluri2021flavr} and VSR\cite{yi2021omniscient,2020BasicVSR} have shown that proper usage of multiple frames can boost restoration performance. However, a model can only process frames of finite length at a time due to GPU memory limitations. From a practical point of view, it is essential to compare the memory consumption and performance of different algorithms with different frame input lengths. To make a fair comparison, we also choose the method that can accept
 multi-frame as input. Specifically, we select Zooming SlowMo\cite{2021Zooming} and TMnet\cite{xu2021temporal}, which adopt the ConvLSTM structure to propagate information within multiple frames. Our experiment is conducted on the
  REDS\cite{Nah_2019_CVPR_Workshops} test set, which consists of 8 sequences, and each sequence contains 100 frames. We downsample the odd-index frames(e.g., $1^{st}$,$3^{rd}$ $\cdots$ )  by a scale factor of 4 and feed them into the models to reconstruct the corresponding HR frames.
Quantitative performance are provided in Table.~\ref{tab:length}. As can be seen, while processing the same number of input frames as its competitors, our approach consumes much less memory. Meanwhile, by increasing the number of input frames, our method can continuously and stably improve the reconstruction performance. We also observed that on some test sequences, the length of the input frames significantly affects the reconstruction
performance. For example, on the Vid4 dataset, if our model accepts four frames as input at a time, then the PSNR is only 26.43. If we directly feed all frames of each sequence into the model, the PNSR will rise to 26.87. This result illustrates the importance of long-distance information for improving the performance of ST-VSR. Please note we conduct experiment on REDS dataset here because  Vid4 dataset contains four sequences with different resolutions
 and lengths, making it difficult to calculate the average memory consumption. These results show that our method can make use of long-distance temporal information more efficiently and effectively. \\

\noindent \textbf{More discussion about performance and efficiency.}
After comparing performance and efficiency, it is necessary to explain why the proposed model is so memory efficient. In general, we fully consider how to save memory when designing the temporal interpolation module and spatial up-sampling module. The specific reasons can be
  \begin{table*}[t]
	\caption{Ablation study (PSNR(dB) / SSIM) on spatial modulation. T $\times$ A , S $\times$ B refers to A temporal interpolated frames  and B space up-sampling  scale. The best performance is highlighted in \textcolor{red}{red}. }
	\label{tab:6}
	\centering
	\resizebox{0.90\linewidth}{!}{\begin{tabular}{c|llllll}
			\hline
			\multicolumn{1}{l|}{} & T $\times$ 2  S $\times$ 2               & T $\times$ 2 S $\times$ 2.4             & T $\times$ 2 S$\times$ 2.8             & T $\times$ 2 S $\times$ 3.2             & T $\times$ 2 S $\times$ 3.6             & T $\times$ 2 S $\times$ 4.0             \\ \hline \hline
			C-STVSR (-b)             & \multicolumn{1}{c}{37.27 / 0.9791} & \multicolumn{1}{c}{28.22 / 0.9018} & \multicolumn{1}{c}{27.21 / 0.8750} & \multicolumn{1}{c}{32.13 / 0.9208} & \multicolumn{1}{c}{26.78 / 0.8477} & \multicolumn{1}{c}{30.30 / 0.8819} \\ \hline
			C-STVSR (-ps1)             & \multicolumn{1}{c}{37.07 / 0.9790} & \multicolumn{1}{c}{34.77 / 0.9650} & \multicolumn{1}{c}{33.95 / 0.9464} & \multicolumn{1}{c}{32.32 / 0.9209} & \multicolumn{1}{c}{\textcolor{red}{30.55} / 0.9018} & \multicolumn{1}{c}{\textcolor{red}{30.36} / 0.8781} \\ \hline
			C-STVSR (-fix)             & \multicolumn{1}{c}{33.54 / 0.9518} & \multicolumn{1}{c}{33.35 / 0.9284} & \multicolumn{1}{c}{33.28 / 0.9337} & \multicolumn{1}{c}{32.55 / 0.9203} & \multicolumn{1}{c}{30.43 / 0.8974} & \multicolumn{1}{c}{30.14 / 0.8861} \\ \hline
			C-STVSR (Ours)                  & \multicolumn{1}{c}{\textcolor{red}{37.65}  / \textcolor{red}{0.9797}}                                 & \multicolumn{1}{c}{\textcolor{red}{34.82} / \textcolor{red}{0.9660}}                                  &\multicolumn{1}{c}{\textcolor{red}{34.19} / \textcolor{red}{0.9495}}                                   & \multicolumn{1}{c}{\textcolor{red}{32.99} / \textcolor{red}{0.9304}}                                  & \multicolumn{1}{c}{\textcolor{red}{30.55} / \textcolor{red}{0.9082}}                                  & \multicolumn{1}{c}{30.24 / \textcolor{red}{0.8865}}                                  \\ \hline
			& T $\times$ 4 S $\times$ 2               & T $\times$ 4 S $\times$ 2.4             & T $\times$ 4 S $\times$ 2.8             & T $\times$ 4 S $\times$ 3.2             & T $\times$ 4 S $\times$ 3.6             & T $\times$ 4 S $\times$ 4.0             \\ \hline
			C-STVSR (-b)             & \multicolumn{1}{c}{36.00 / 0.9707} & \multicolumn{1}{c}{28.08 / 0.8966} & \multicolumn{1}{c}{27.14 / 0.8701} & \multicolumn{1}{c}{31.48 / 0.9106} & \multicolumn{1}{c}{26.70 / 0.8420} & \multicolumn{1}{c}{\textcolor{red}{29.88} / 0.8718} \\ \hline
			C-STVSR (-ps1)             & \multicolumn{1}{c}{35.82 / 0.9706} & \multicolumn{1}{c}{33.81 / 0.9557} & \multicolumn{1}{c}{33.02 / 0.9355} & \multicolumn{1}{c}{31.59 / 0.9099} & \multicolumn{1}{c}{30.05 / 0.8904} & \multicolumn{1}{c}{29.85 / 0.8674} \\ \hline
			C-STVSR (-fix)             & \multicolumn{1}{c}{32.85 / 0.9436} & \multicolumn{1}{c}{32.58 / 0.9191} & \multicolumn{1}{c}{32.42 / 0.9222} & \multicolumn{1}{c}{31.72 / 0.9081} & \multicolumn{1}{c}{29.92 / 0.8853} & \multicolumn{1}{c}{29.72 / 0.8751} \\ \hline
			C-STVSR (Ours)                  &  \multicolumn{1}{c}{\textcolor{red}{36.29} / \textcolor{red}{0.9717}}                                & \multicolumn{1}{c}{\textcolor{red}{33.89} / \textcolor{red}{0.9574}}                                 & \multicolumn{1}{c}{\textcolor{red}{33.27} / \textcolor{red}{0.9393}}                                 & \multicolumn{1}{c}{\textcolor{red}{32.22} / \textcolor{red}{0.9194}}                                 &\multicolumn{1}{c}{\textcolor{red}{ 30.08} / \textcolor{red}{0.8967}}                                   &       \multicolumn{1}{c}{29.78 / \textcolor{red}{ 0.8758}}                           \\ \hline
	\end{tabular}}
	
\end{table*}
  summarized as follows:  First, the motion information is estimated and used in a very efficient manner. The optical flow is estimated only once but used three times. Specifically, we extract the optical flow using Spynet\cite{ranjan2017optical}, which is first used to align features in the first stage. Next, the optical flow is used again in the second phase to synthesize the 
\begin{figure}[t]
	\centering
	\includegraphics[width=8cm]{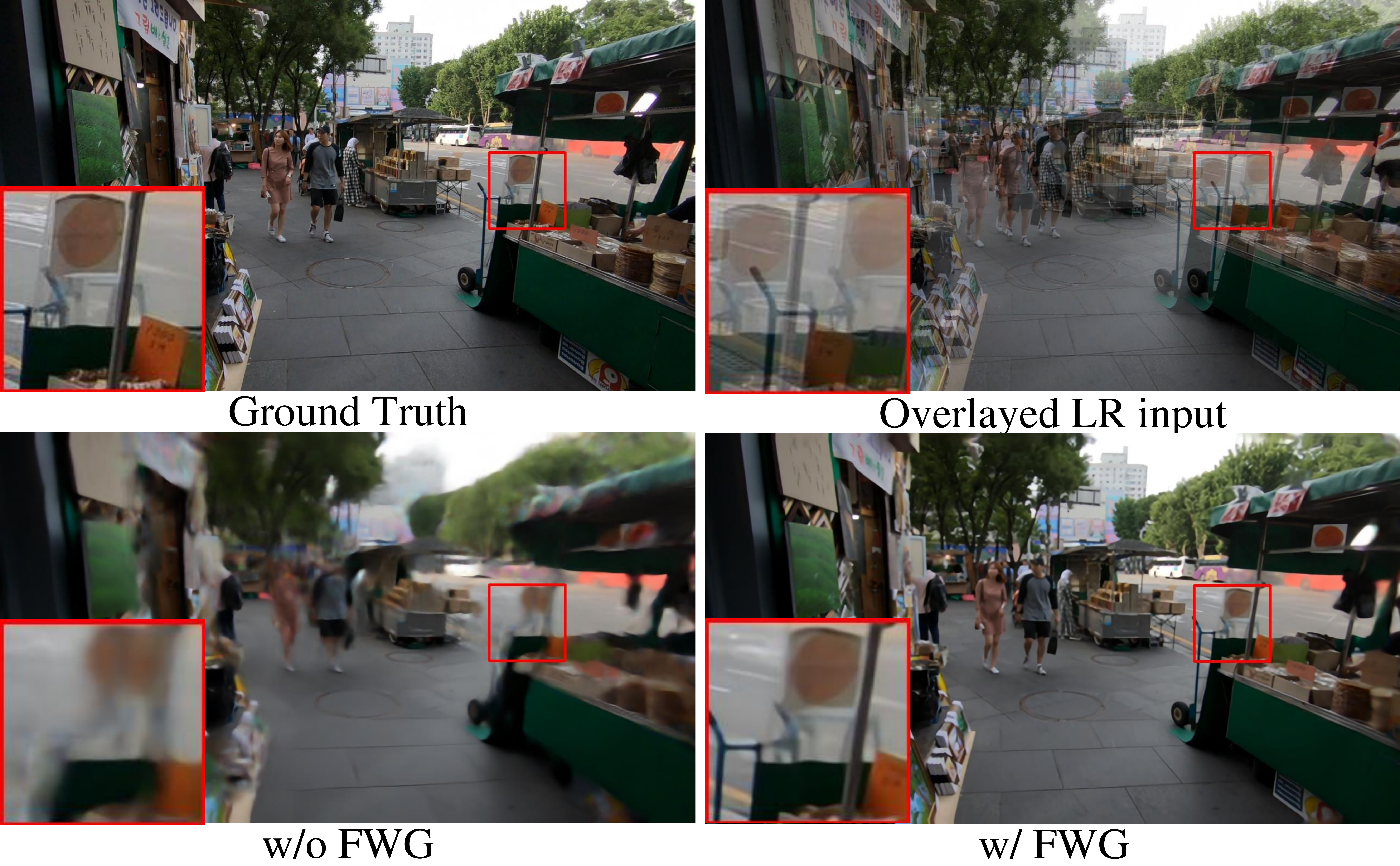}
	\caption{Visual comparison of with or without forward warping guidance.  Part of the areas are zoomed in for better view } \label{fig:8}
\end{figure} 
 intermediate temporal frames. Moreover, the optical flow is finally used as guidance information to generate pseudo-labels for training supervision. In contrast, other RNN-based methods\cite{2021Zooming,xu2021temporal} estimate the motion information of the pre-existing frames and temporally interpolated frames separately, which ignores the spatio-temporal connection between the VFI and the VSR task. Secondly, unlike other complex and elaborate upsampling module designs, we have designed a very straightforward cascading pixel shuffle module. This kind of depth-to-space operation itself does not introduce any learnable parameters. We opt not to design complex conditional
convolutions\cite{hu2019meta,wang2021learning} or implicit local representation\cite{chen2022videoinr} to accurately estimate the  weight of each spatial location. Instead, we aim to design a memory-friendly module that can efficiently exploit long-range information.
\subsection{Ablation Study}
To further investigate the proposed algorithm, we design a set of ablation experiments for different modules. 
\subsubsection{temporal modulation}
In this section, we demonstrate the effect of each component that contributes to temporal modulation. The quantitative results are shown in Table~\ref{tab:5}.\\
\textbf{Forward warping guidance (FWG)}
\begin{figure}[t]
	\centering
	\includegraphics[width=8cm]{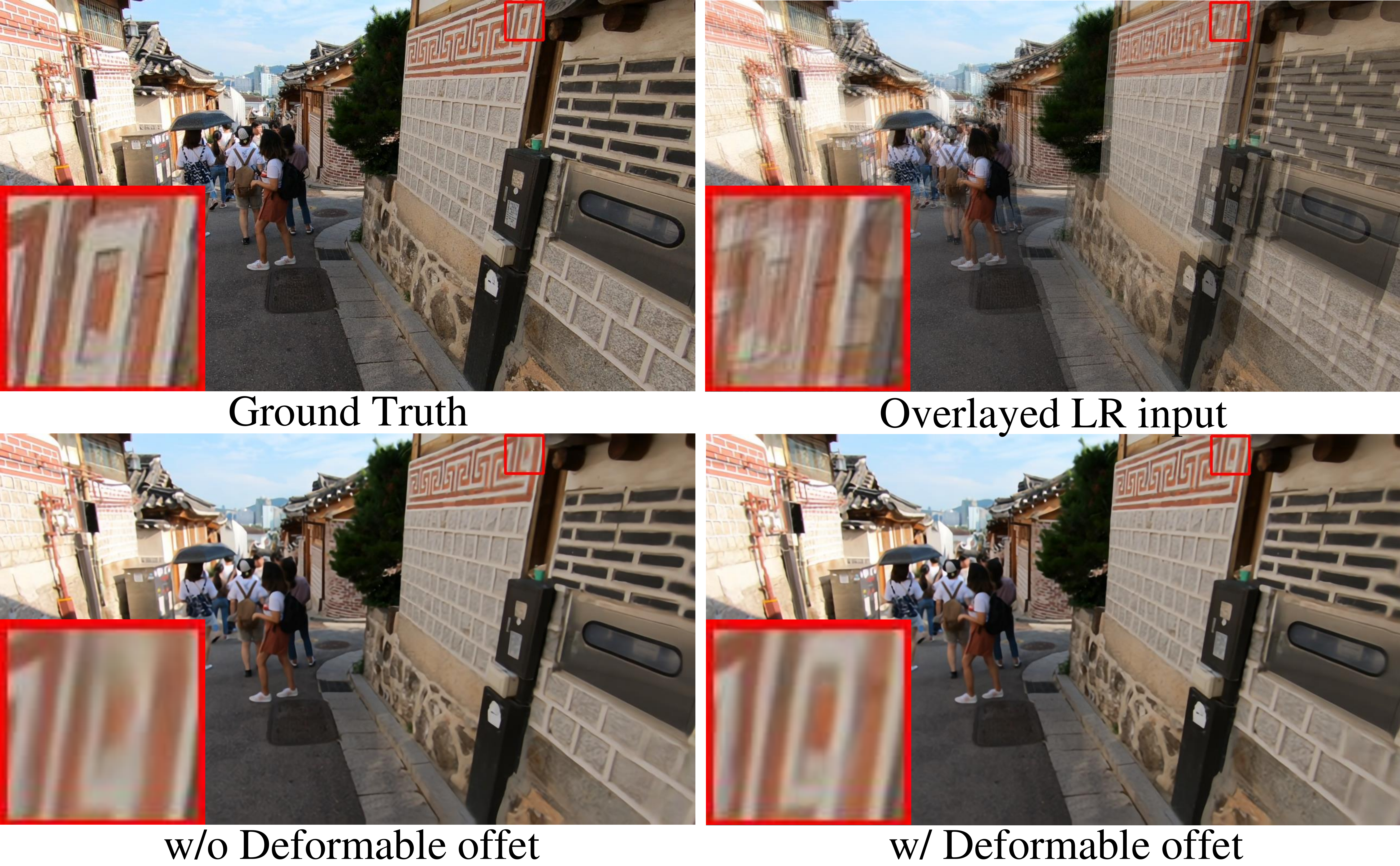}
	\caption{Visual comparison of with or without deformable alignment. Part of the  areas are zoomed in for better view.}\label{fig:9}
\end{figure}
We compare two alignment methods: using only deformable convolution to approximate the intermediate frames and using forward warping to
guide the deformable alignment.
As seen in Table~\ref{tab:5}, the absence of FWG results in a drop in performance on Vimeo and REDS datasets. This suggests that forward warping can help synthesize more reliable intermediate frames. We also observe that forward warping guidance may considerably influence subjective visual quality in some cases, especially when approximating large motions. A typical example is shown in Fig.~\ref{fig:8}, there exists extremely large motion between two consecutive frames. The model with FWG shows overwhelming subjective visual improvements. We must emphasize that sometimes objective indicators(PSNR or SSIM) cannot reflect the subjective visual perception, especially for the video frame interpolation task. Although our method synthesizes visually pleasing results, there is still a considerable positional bias between the predicted frame and the ground truth due to the fact that much of the real-world motion is non-linear. \\
\textbf{Deformable alignment}
 Although forward warping can already synthesize visually appealing results, using only off-the-shelf optical flow estimators may still not be flexible enough. This is because optical flow estimation is not completely accurate. Therefore, we have further introduced deformable alignment(DA). We expect the model can adaptively learn various complex motions during training. The objective results on whether to add deformable convolution are provided in Table.~\ref{tab:5}. It can be seen that deformable alignment after forward warping can further improve the restoration performance. Visual comparisons are illustrated in Fig.~\ref{fig:9}. We found that deformable alignment can help to restore sharper details.\\
\textbf{Flow-guided texture consistency loss}
 \begin{figure}[t]
	\centering
	\includegraphics[width=8cm]{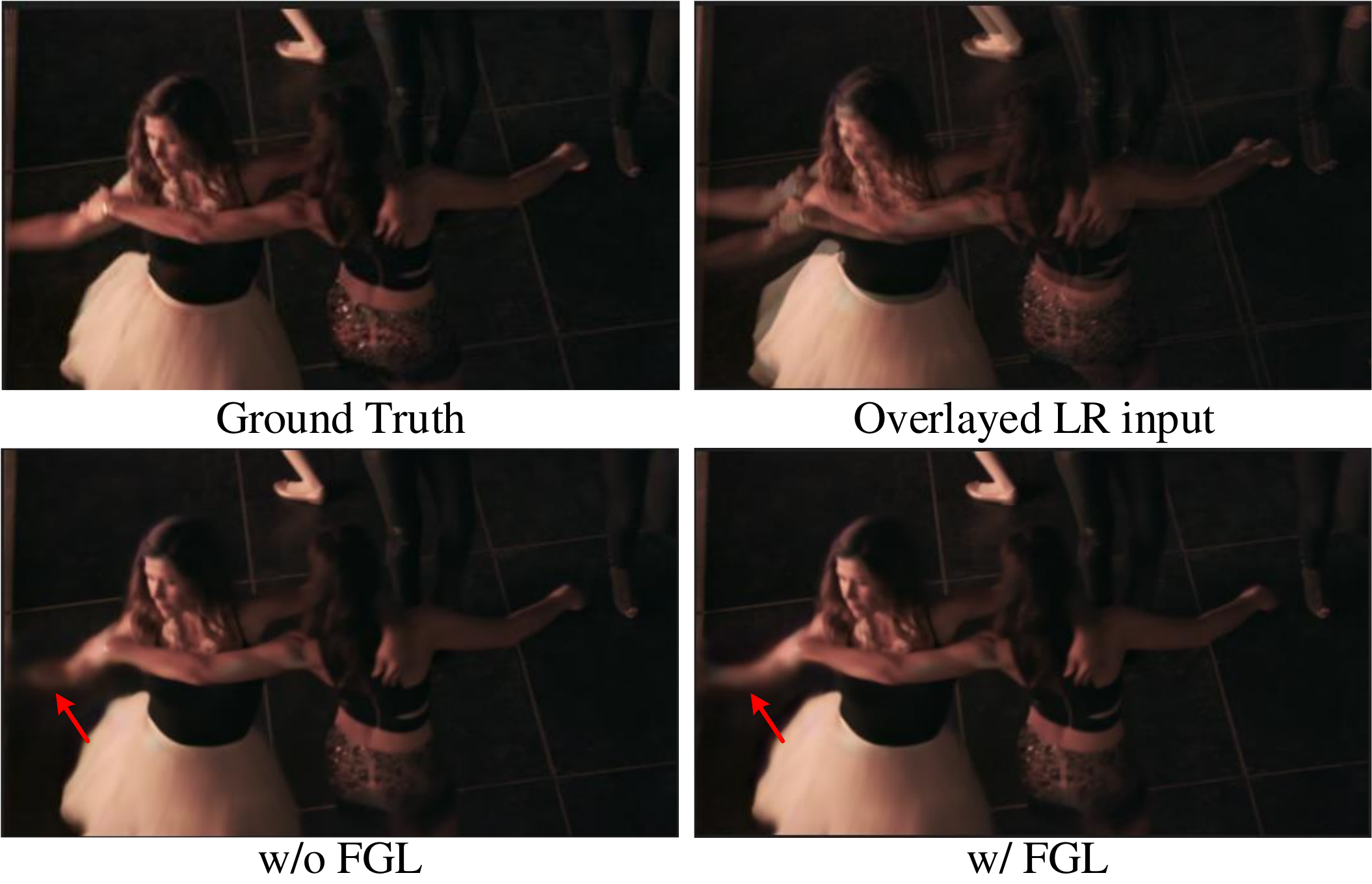}
	\caption{ Visual comparison of results for with or without flow-guided texture consistency loss. Pay attention to the areas indicated by red arrows, zoom in for a better view.} \label{fig:11}
\end{figure}
Based on Zhou \emph{et al.}'s work\cite{zhou2022exploring}, we propose a flow-guided texture consistency loss (FGL) to relax the strict restrictions on ground truth of intermediate frames and greatly reduce its searching space. Through experiments, we observe that this loss also improves reconstruction performance and helps preserve the structures and patterns of interpolated contents. We provide a visual comparison in Fig.~\ref{fig:11}. 
A dancer's arm is moving from the top to down. Note the positions indicated by the red arrows, FGL can help restore a complete semantic pattern.
\subsubsection{spatial modulation}
We next focus on the spatial upsampling component, including model design and training strategies. All these models are trained on Vimeo90K-T\cite{DBLP:journals/corr/abs-1711-09078}  and tested on the SPMCS\cite{tao2017detail} dataset. The comparison results are shown in Table~\ref{tab:6}.\\
\textbf{Module design}
To verify the validity of the proposed model, we first establish a baseline with the bilinear operation. To be specific, we simply replace stage 3 with a bilinear function followed by a 1$\times$1 Conv to accommodate different scales. We denote this baseline model as C-STVSR (-b), as can be seen in Table~\ref{tab:6}, directly scaling the features with bilinear interpolation results in severe performance fluctuations. 
We also observe that such an upsampling approach may do harm to training stability. This is probably due to the direct interpolation of the generated feature maps ignoring the scale information. But the scale information is essential for scale-arbitrary super-resolution.
We then study the cascading upsampling module. To verify the validity of the multi-scale depth-to-space module, we trained a model which only contains a single pixel shuffle layer( 4$\times$ spatial magnification) and denoted it as C-STVSR (-ps1). As can be seen in Table~\ref{tab:6}, it slightly suffers in most situations when compared with the muti-scale depth-to-space module, especially for the missing scales that exist in the cascading module( 2$\times$ in this case). Since depth-to-space layers bring almost no additional compute costs, we sampled the multi-scale features to further enhance the model's adaptability to accommodate different scales.\\
 \textbf{Training stratgedy.}
 As noted above, we train our model with different scales to force the model to learn scale-aware information. To explore the effectiveness of different training strategies, we train the model with a fixed scale (space 4 $\times$ upscaling) and test it on different scales. When the test size is close to the training size, the model's performance is comparable to C-STVST, but when the test scale varies, there is a severe drop in performance. This suggests that training data at different scales has a huge effect on the network's ability to adapt to different super-resolution magnifications.

\section{Conclusions}

\label{sec:concl}
This paper presents a flexible and accurate structure for space-time video super-resolution, which can modulate the given video to arbitrary frame rate and spatial resolution. We first perform multi-frame information enhancement with a bi-directional RNN structure. After that, a memory-friendly forward warping guided feature alignment module is proposed to synthesize intermediate frames at an arbitrary intermediate time. To better preserve texture consistency in the intermediate reconstructed frames, we propose an optical flow guided pseudo-label generation strategy that greatly optimizes the search space. Further, all motion information is organized in an efficient manner, which greatly avoids the redundant estimation of motion information. Finally, a straightforward yet efficient cascaded upsampling module is proposed, which enables our model to achieve scale-arbitrary upsampling and meanwhile, significantly reduce memory consumption. With the incorporation of these techniques into an end-to-end framework, our model can make usage of long-range information with high efficiency. Extensive experiments have demonstrated that our approach not only has good flexibility but also achieves significant performance gains over existing methods.

%
%

\bibliographystyle{IEEEtran}
\bibliography{IEEEabrv,./bibliography}
\end{document}